\pdfoutput=1
\documentclass[11pt]{article}

\usepackage[final]{acl}

\usepackage{times}
\usepackage{latexsym}

\usepackage[T1]{fontenc}
\usepackage[utf8]{inputenc}

\usepackage{microtype}

\usepackage{inconsolata}
\usepackage{array}
\usepackage{tabularx}
\usepackage{amsmath,amsfonts,bm}

\def\eqref#1{equation~\ref{#1}}
\def\1{\bm{1}}

\DeclareMathAlphabet{\mathsfit}{\encodingdefault}{\sfdefault}{m}{sl}
\SetMathAlphabet{\mathsfit}{bold}{\encodingdefault}{\sfdefault}{bx}{n}

\usepackage{color,xcolor}
\usepackage{epsfig}
\usepackage{graphicx}

\usepackage{adjustbox}
\usepackage{array}
\usepackage{booktabs}
\usepackage{colortbl}
\usepackage{wrapfig}
\usepackage{hhline}
\usepackage{multirow}
\usepackage{subcaption} % issues a warning with CVPR/ICCV format
\usepackage{wrapfig}
\usepackage{floatflt}

\usepackage{amsmath,amsfonts,amssymb}
\usepackage{mathtools}  % additional useful math commands, such as \coloneqq
\usepackage{bm}
\usepackage{nicefrac}
\usepackage{changepage}
\usepackage{extramarks}
\usepackage{fancyhdr}
\usepackage{lastpage}
\usepackage{setspace}
\usepackage{soul}
\usepackage{xspace}

\usepackage{url}
\usepackage{enumitem}  % control indent of enumerate, itemize, etc.

\usepackage{makecell}

\usepackage{pifont} % for extra symbols such as \cmark

\usepackage{algorithm}
\usepackage[noend]{algpseudocode}

\algnewcommand\algorithmicswitch{\textbf{switch}}
\algnewcommand\algorithmiccase{\textbf{case}}
\algnewcommand\algorithmicassert{\textbf{assert}}
\algnewcommand\algorithmiclet{\textbf{let}}
\algnewcommand\Assert[1]{\State \algorithmicassert\xspace #1}%
\algnewcommand\Let[1]{\State \algorithmiclet\xspace #1}%
\algnewcommand\Yield{\textbf{yield}\xspace}%
\algnewcommand\Break{\textbf{break}\xspace}%
\algnewcommand\YieldFrom{\textbf{yield from}\xspace}%
\algdef{SE}[SWITCH]{Switch}{EndSwitch}[1]{\algorithmicswitch\ #1\ \algorithmicdo}{\algorithmicend\ \algorithmicswitch}%
\algdef{SE}[CASE]{Case}{EndCase}[1]{\algorithmiccase\ #1}{\algorithmicend\ \algorithmiccase}%
\algtext*{EndSwitch}%
\algtext*{EndCase}%

\usepackage{framed}
\definecolor{shadecolor}{gray}{0.95}

\usepackage{marginnote}

\newcolumntype{L}[1]{>{\raggedright\let\newline\\\arraybackslash\hspace{0pt}}m{#1}}
\newcolumntype{C}[1]{>{\centering\let\newline\\\arraybackslash\hspace{0pt}}m{#1}}
\newcolumntype{R}[1]{>{\raggedleft\let\newline\\\arraybackslash\hspace{0pt}}m{#1}}

\usepackage{pifont}% http://ctan.org/pkg/pifont
\usepackage{float}

\usepackage[many]{tcolorbox}    	% for COLORED BOXES (tikz and xcolor included)
\newtcolorbox{boxA}{
    sharpish corners,
    colback = lightgray!40, % background color
    boxrule = 0pt  % no borders
}

\newcommand{\sect}[1]{Section~\ref{#1}}
\newcommand{\sectapp}[1]{Appendix~\ref{#1}}

\newcommand{\fig}[1]{Fig.~\ref{#1}}
\newcommand{\tbl}[1]{Table~\ref{#1}}

\newcommand{\ignore}[1]{}

\makeatletter
\DeclareRobustCommand\onedot{\futurelet\@let@token\@onedot}
\def\@onedot{\ifx\@let@token.\else.\null\fi\xspace}

\def\eg{e.g\onedot} 
\def\ie{i.e\onedot}

\def\etal{et al\onedot}

\makeatother

\definecolor{MyDarkBlue}{rgb}{0,0.08,1}
\definecolor{MyDarkGreen}{rgb}{0.02,0.6,0.02}
\definecolor{MyDarkRed}{rgb}{0.8,0.02,0.02}
\definecolor{MyDarkOrange}{rgb}{0.40,0.2,0.02}
\definecolor{MyPurple}{RGB}{111,0,255}
\definecolor{MyRed}{rgb}{1.0,0.0,0.0}
\definecolor{MyGold}{rgb}{0.75,0.6,0.12}
\definecolor{MyDarkgray}{rgb}{0.66, 0.66, 0.66}
\definecolor{JiayuanColor}{rgb}{0.60,0.43,0.48}
\definecolor{XinyiColor}{rgb}{0.37,0.62,0.63}

\newcommand{\model}{DetailScribe\xspace}
\newcommand{\data}{InterActing\xspace}

\newif\ifpropositionfirstitem
\propositionfirstitemtrue

\newcommand{\myparagraph}[1]{\noindent\textbf{#1}}
\newcommand{\xhdr}[1]{\noindent\textbf{#1}}

\newcommand{\mycell}[1]{\begin{tabular}[t]{@{}l@{}l}#1\end{tabular}}

\usepackage{enumitem}
\setitemize{itemsep=1pt,topsep=1pt,parsep=0pt,partopsep=0pt, leftmargin=*}
\title{Generating Fine Details of Entity Interactions}

\author{Xinyi Gu \and Jiayuan Mao \\
        Massachusetts Institute of Technology\\
  \texttt{\{gxy, jiayuanm\}@mit.edu}}

\begin{document}
% \maketitle

\twocolumn[{%
\renewcommand\twocolumn[1][]{#1}%
% \vspace{-1em}

\maketitle
\vspace{-1.2em}
\includegraphics[width=\textwidth]{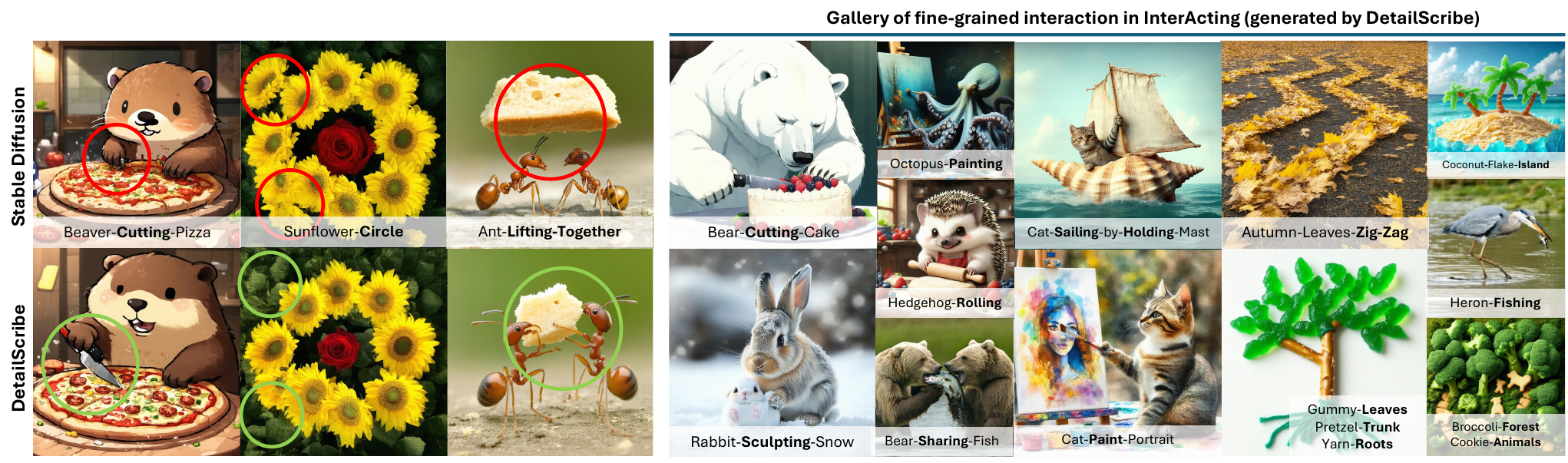}
\vspace{-1.5em}
\captionof{figure}{Left: DetailScribe improves the base text-to-image model across three scenarios: functional interaction, complex scene layouts, and multi-subject interactions. Right: A gallery showcasing \model-generated images with rich entity interactions.}
\label{fig:teaser}
\vspace{1.2em}
}]

\begin{abstract}
Recent text-to-image models excel at generating high-quality object-centric images from instructions. However, images should also encapsulate rich interactions between objects, where existing models often fall short, likely due to limited training data and benchmarks for rare interactions. This paper explores a novel application of Multimodal Large Language Models (MLLMs) to benchmark and enhance the generation of interaction-rich images.
We introduce \data, an interaction-focused dataset with 1000 LLM-generated fine-grained prompts for image generation covering (1) functional and action-based interactions, (2) multi-subject interactions, and (3) compositional spatial relationships.
To address interaction-rich generation challenges, we propose a decomposition-augmented refinement procedure. Our approach, \model, leverages LLMs to decompose interactions into finer-grained concepts, uses an MLLM to critique generated images, and applies targeted refinements with a partial diffusion denoising process. Automatic and human evaluations show significantly improved image quality, demonstrating the potential of enhanced inference strategies. Our dataset and code are available at \url{https://detailscribe.github.io/}.

\end{abstract}
\section{Introduction}
\label{sec:intro}

Recent advances in text-to-image (T2I) generation have enabled models to create highly realistic images that capture a diverse set of objects with varying attributes, colors, and textures from natural language descriptions. However, while these models excel at generating individual objects or simple scenes, they often struggle when tasked with producing images that involve intricate interactions between entities or complex spatial layouts. The challenge becomes particularly pronounced when the interactions are uncommon or abstract, and when the subjects involved deviate from familiar, human-centered scenarios. For instance, generating scenes that depict animals using tools, rather than humans, or rendering abstract structures like mazes with precise spatial arrangements poses a significant challenge.
A key limitation is the absence of datasets designed for training and evaluating complex interactions.

\begin{table*}[tp]
    \centering\small
    \setlength{\tabcolsep}{3pt}
    % \setlength{\extrarowheight}{4pt}
    % \begin{tabular}{lcc}
    \begin{tabular}{>{\raggedright\arraybackslash}m{0.28\linewidth}m{0.20\linewidth}m{0.48\linewidth}}
    % \begin{tabular}{lll}
    \toprule
         \textbf{Scenario} & \textbf{Subclass} &  \textbf{Examples} \\
         \midrule
         \multirow{2}{*}{\mycell{\bf Functional and\\ \bf Action-Based Interactions (600)}} 
         & Tool Manipulation (227) & cutting, painting, sailing, stirring, taking a photo \\
        \cmidrule{2-3}
        & Physical Contact (373)  & sculpting snow, stacking, holding  \\
        \midrule
    % \multirow{2}{*}{\mycell{\bf Multi-subject \\ \bf Interactions}} 
    % & \parbox[c][2cm][c]{\linewidth}{Interaction}  & huddling, high-five, collaborating to lift, weaving leaves together, sharing food \\
         \textbf{Multi-subject Interactions (200)}  & Interaction (200) & huddling, high-five, collaborating to lift, weaving leaves together, sharing food  \\
        \midrule
         \multirow{2}{*}{\mycell{\bf Compositional \\ \bf Spatial Relationships (200)}}  &  Abstract Layouts (183) & tic-tac-toe, table, atom, solar system, forest, tree, bookshelf \\
         \cmidrule{2-3}
         & Geometric Patterns (17) & zig-zag pattern, circle, center \\
         \bottomrule
    \end{tabular}
    \vspace{-0.5em}
    \caption{The \data dataset contains 1000 text-to-image prompts by subclass and occurrence.}
    \vspace{-1em}
    \label{tab:datastats}
\end{table*}

In particular, many existing benchmarks for text-to-image models have been focusing on single objects or simple spatial relations. To address this, we propose a new Large language models (LLMs) generated dataset specifically curated for fine-grained and interaction-rich text-to-image generation. The dataset includes examples of functional and action-related interactions (\eg, using tools and making physical contacts), multi-subject interactions, and compositional relationships (\eg, geometric and abstract layouts). We evaluate models using a combination of MLLM assessments, automatic metrics, and human evaluation protocols for a more comprehensive and robust measure of generation quality.
\fig{fig:teaser} shows common failures of off-the-shelf T2I models~\citep{esser2024scalingrectifiedflowtransformers} such as physically-feasible interactions and layout errors.

To enhance fine-grained interaction generation in T2I models, we introduce {\it \model}, a generate-then-refine framework based on {\it concept decomposition}.
\model is the first framework to combine multi-modal LLMs’ reasoning (concept decomposition) and recognition (image critique) ability to improve text-to-image generation. It is compatible with most T2I models, preserves their diversity, and requires no additional datasets or domain-specific knowledge.
At a high level, \model has two steps: generating an initial image from a prompt, followed by iterative refinement based on MLLM critiques.

We compare our approach with the state-of-the-art text-to-image generation frameworks, and demonstrate that our concept-decomposition-based refinement significantly improves generation quality across a range of challenging scenarios. In summary, our contributions are: (1) We propose the \data dataset for text-to-image generation with fine-grained interactions. (2) We benchmark several previous text-to-image models on \data and propose a new framework, \model, to improve T2I generation by integrating multi-modal LLMs for reasoning and recognition. It features a structured decomposition approach to enhance critique-based refinement. (3) Our experiments demonstrate that DetailScribe improves generation quality across a variety of challenging scenarios.

\section{Related Works}
\label{sec:related}

\begin{figure*}[tp]
   \centering
   \includegraphics[width=1\linewidth]{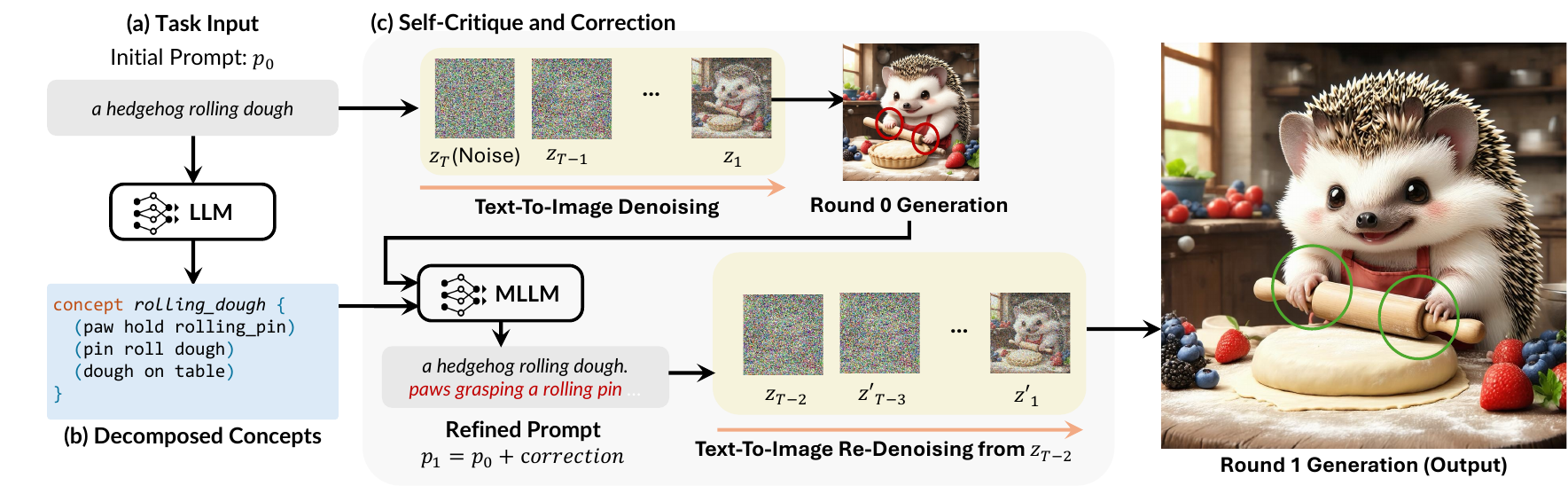}
   \caption{The overall pipeline of \model. \model takes as input a single natural language instruction. It first prompts a large language model (LLM) to generate a breakdown of the concepts in the image, which guides a multimodal LLM (MLLM) to attend to different regions of a generated image and suggests fixes. It then adds noises back to the generated image and re-runs the diffusion process with the MLLM-refined prompt.}
   \label{fig:pipeline}
   \vspace{-1em}
\end{figure*}

% \subsection{Text-to-Image Diffusion Models}
\myparagraph{Text-to-image diffusion models.}
Diffusion models have significantly advanced the field of image generation, especially image fidelity~\citep{ho2020denoising,ramesh2022hierarchicaltextconditionalimagegeneration,saharia2022photorealistic,ho2022imagenvideohighdefinition,esser2024scalingrectifiedflowtransformers,imagenteamgoogle2024imagen3}. However, challenges remain in handling complex relationships and intricate compositional structures. Another approach to improving fine details of object interactions is by adding additional images at inference time (\ie, customization~\citealp{huang2023reversiondiffusionbasedrelationinversion,huang2024learningdisentangledidentifiersactioncustomized}). Yet their approach is orthogonal to ours because we do not rely on any user-provided images.

% \subsection{Text-to-Image Benchmarks}
\myparagraph{Text-to-image benchmarks.}
% \vspace{-0.25em}
Most of the existing image-generation benchmarks focus on automated metrics such as Inception Score~\citep{salimans2016inceptionscore}, FID~\citep{heusel2017fid}, and CLIPScore~\citep{hessel2022clipscorereferencefreeevaluationmetric} to assess image quality and alignment on image datasets such as MS-COCO~\citep{lin2014mscoco} and ImageNet~\citep{deng2009imagenet}. Some human preference \citep{liang2024richhumanfeedbacktexttoimage} studies have been conducted by requesting users to rank and rate images~\citep{xu2023imagerewardlearningevaluatinghuman,kirstain2023pickapicopendatasetuser}. Recently, more comprehensive synthetic benchmark datasets have been proposed for holistic evaluation~\citep{lee2023holisticevaluationtexttoimagemodels, huang2023t2icompbenchcomprehensivebenchmarkopenworld}.
% and proposed a comprehensive compositional text-to-image (T2I) generation benchmark, consisting of text prompts from color binding, shape binding, texture binding, spatial relationships, non-spatial relationships, and complex compositions.
Nevertheless, most existing works focus on the generation of detailed attributes that are explicitly described in the prompt, while still struggling to accurately modeling interactions among objects. In this work, we craft \data to evaluate such ability of T2I models.

% \subsection{Inference Scaling and Self-Correction}
\myparagraph{Inference scaling and self-correction.}
% \vspace{-0.25em}
MLLMs have shown potential for evaluating the performance of other models or even themselves. During training, MLLMs are often used to provide feedback rewards that improve alignment~\citet{rafailov2024direct}. During inference, self-correction mechanisms have also been applied for improved performance~\citep{pan2023automatically, zhang2023summit, gao2023rarr, dong2024self, alphaproof2024}. Most recently, ~\citet{ma2025inferencetimescalingdiffusionmodels} introduces an inference-time scaling framework for diffusion models by searching for optimal noise inputs. Recent advancements have incorporated MLLMs as agents to control the generation of diffusion models~\citep{zhang2023controllabletexttoimagegenerationgpt4,lian2024llmgroundeddiffusionenhancingprompt,feng2023layoutgptcompositionalvisualplanning, wu2024self, cho2023visualprogrammingtexttoimagegeneration, wang2024genartistmultimodalllmagent}. Most works still rely on explicit object grounding and struggle to identify the fine-grained sub-part interactions. Our method focuses on more granular feedback and attempting to retain partial diffusion steps for better performance.
\section{The \data Dataset}

We proposed a new dataset generated by LLM, {\it \data} consisting of 1000 interaction-focused text prompts from 3 scenarios covering three major types or real-world interactions: functional relationships and action-based interactions (600), including tool manipulation and actions that involve rich physical contact, multi-subject interactions (200), and compositional spatial relationships (200), which are usually presented in the form of objects forming abstract layouts of geometric patterns. In contrast to existing efforts on text-to-image benchmarking focusing on single object generation~\citep{wu2024self}, combination of spatial and attribute relationships~\citep{huang2023t2icompbenchcomprehensivebenchmarkopenworld}, and holistic aspects such as aesthetics and multi-linguality~\citep{lee2023holisticevaluationtexttoimagemodels}, \data focuses on tasks involving entity interaction with non-trivial details.
We include more examples and statistics of the entire dataset in Appendix~\ref{sec:app:dataset}.

\subsection{Evaluation Metrics}
% \todo{move metrics to experiments}
Due to the challenges in assessing whether an image aligns with the prompt's description, we primarily rely on human evaluation, referred to as the \textbf{human Likert scale}. Human annotators were instructed to assign a score between 1 and 5 based on image-text alignment. Detailed evaluation guidelines are listed Appendix C.1. We further explored the use of MLLMs and pre-trained metrics for automatic evaluation purposes (\textbf{automatic evaluation}). We note that these evaluations are inherently noisier, so we compared their agreement with human preferences on sampled image pairs generated by all models. We then use the most aligned scores and MLLM questions as our auto-evaluators for benchmarking on the entire \data.

\paragraph{Automatic evaluation.}
We also leverage the reasoning abilities of MLLMs for evaluation. The MLLM evaluation process uses a prompt that includes the rating instruction, and the image generated by a model (Appendix C.2). The MLLM evaluator then outputs a score on a scale of 1 to 5. Besides MLLM, we adopted CLIPScore \cite{hessel2022clipscorereferencefreeevaluationmetric} and ImageReward \cite{xu2023imagerewardlearningevaluatinghuman} as pre-tained text-image matching metrics to assess the image-text alignment of the generation. We further include BLIP-VQA proposed in \cite{huang2023t2icompbenchcomprehensivebenchmarkopenworld} to capture fine-grained text-image correspondences.

\begin{figure*}[tp]
\centering
\includegraphics[width=1\linewidth]{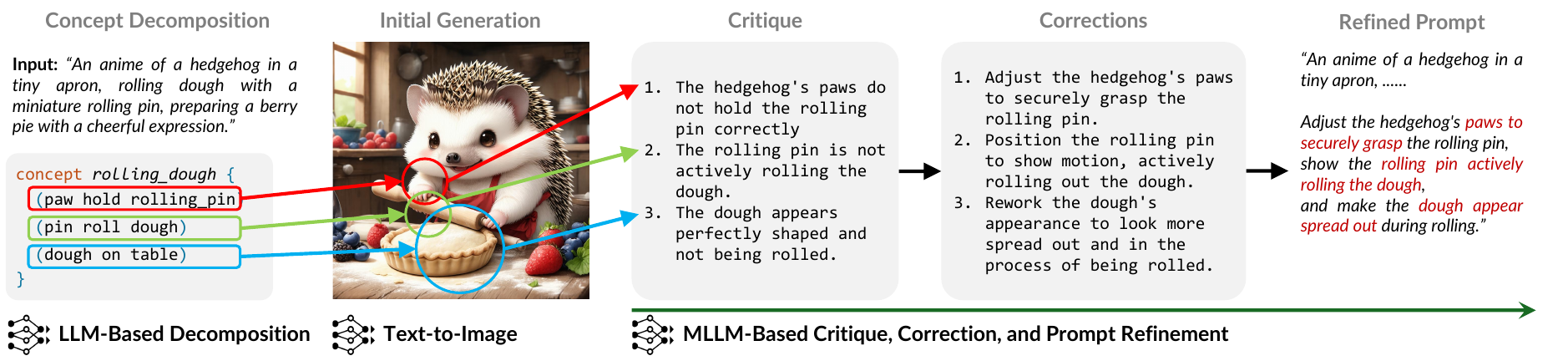}
\caption{MLLM-based critique and prompt refinement. Given the LLM-generated concept decomposition and an image generated using the user input, an MLLM generates a critique of errors in the image, suggests corrections, and finally refines the prompt. This prompt will be used in a second-round diffusion process.}
\label{fig:refine}
\vspace{-1em}
\end{figure*}
\section{\model}

In this section, we introduce our refinement-augmented generation framework ({\it \model}) for text-to-image generation modeling fine-grained entity interactions. Illustrated in \fig{fig:pipeline}, \model operates in three stages: 1) given an input natural language prompt, a large language model hierarchically decomposes it into detailed sub-concepts (\sect{ssec:hierarchical-decomposition}); 2) an initial image is generated from the prompt using a text-to-image model, followed by an MLLM critique conditioned on both the decomposed sub-concepts and the generated image (\sect{ssec:critique}); 3) based on the critique, the prompt is refined and a re-denoising process corrects errors, yielding a more faithful and realistic generated image (\sect{ssec:finetune}).

Overall, our framework leverages a diffusion-based text-to-image model as the base model for generating and refining images. Augmenting this base model, \model is designed to handle highly variable user inputs that describe complex interactions between entities.

\subsection{Prompt Completion by Concept Decomposition}
\label{ssec:hierarchical-decomposition}

At the core of our pipeline, this module refines a user-provided natural language instruction by generating a more detailed, structured version of it. Specifically, we adopt the concept of visual abstraction schema proposed by Hsu \etal~\cite{hsu2024maze}, which represents scene structures with a directed acyclic graph (DAG). Illustrated in \fig{fig:pipeline}, in a schema, each node represents a subcomponent of the higher-level concept, and dependencies between components are captured as edges. For example, the instruction ``a hedgehog is rolling dough with a rolling pin'' can be decomposed into distinct entities (e.g., the hedgehog, the dough, and the rolling pin) and their interactions (e.g., the hedgehog holding the rolling pin, and the rolling pin contacting the dough). Optionally, background elements like tables and windows can be included.

This relational and hierarchical representation is highly flexible, and we find that explicitly performing this decomposition enhances the downstream MLLMs’ ability to capture fine-grained entity interactions, as it naturally provides a ``checklist'' for identifying errors and refining the prompt.

Across all examples shown in this paper, we prompt LLM with only a {\it single} in-context learning example (cooking). We provide details about our text prompts in Appendix A.

\begin{figure*}
   \centering
   \includegraphics[width=.98\linewidth]{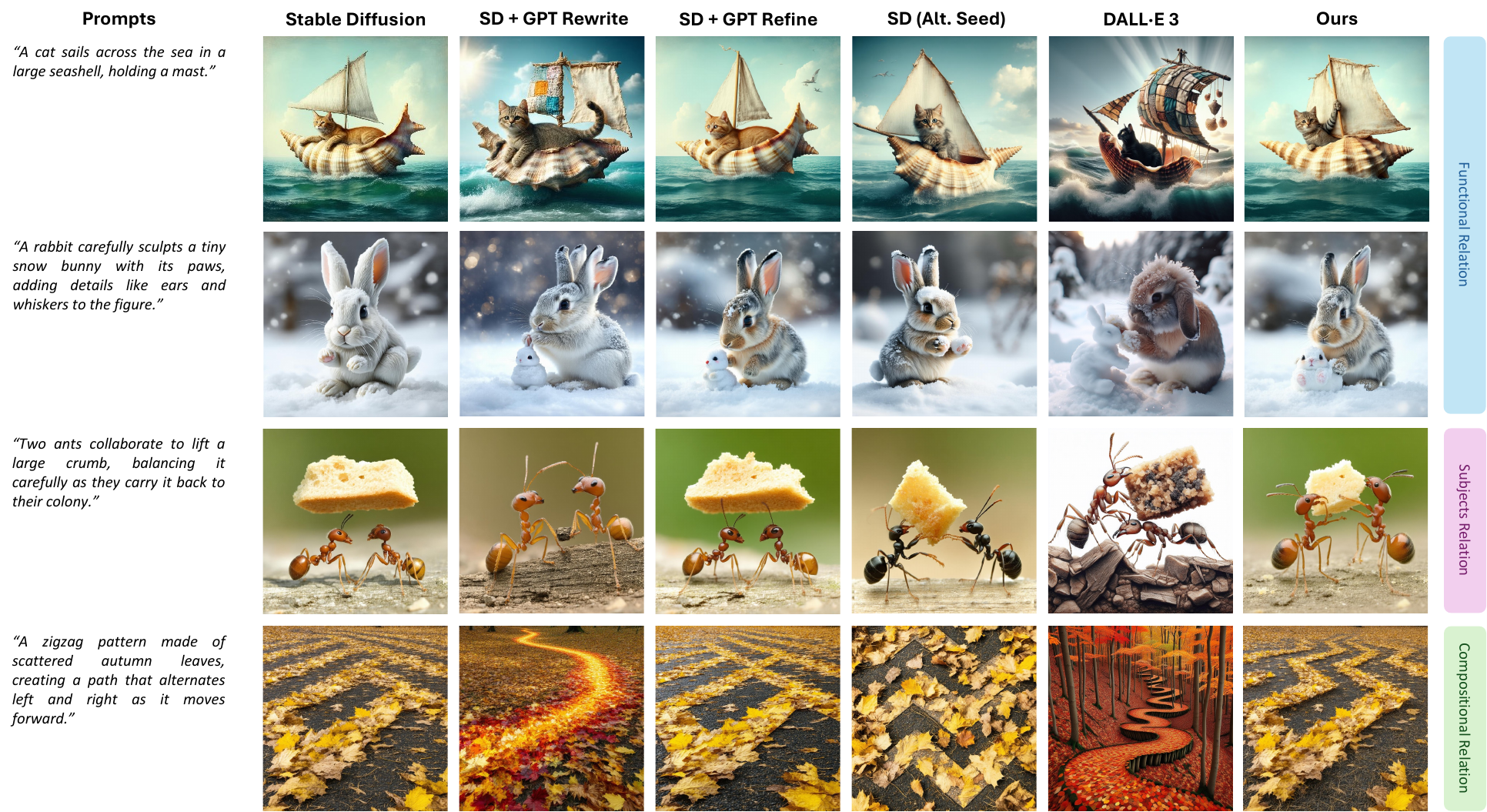}
   \caption{Images generated by \model and baselines on the \data dataset. From left to right: 1) Stable Diffusion; 2) SD + GPT Rewrite; 3) SD + GPT Refine; 4) SD by Alternative seed/initial noise in the Inference scaling experiment; 5) DALL·E 3; 6) Ours: \model. \model consistently provides effective corrections, which help generate images that closely follow the fine details in the prompts.}
   \label{fig:benchmark}
   \vspace{-1em}
\end{figure*}

\subsection{Multimodal LLM-Based Critique and Prompt Refinement}
\label{ssec:critique}

At a high level, this module leverages a multimodal LLM (MLLM) to refine the generated image iteratively. Initially, an image is generated using a base T2I model conditioned on the original user input. Next, we utilize an MLLM (GPT-4o in our experiments) to critique the generated image. The MLLM is prompted to review each element in the schema from \sect{ssec:hierarchical-decomposition}, checking whether the generated image accurately reflects the specified entities and their interactions.

After the MLLM identifies discrepancies, it proposes edits to the original prompt—typically by inserting a few focused phrases to enhance clarity or detail in specific parts of the sentence. This refined prompt is then used in a subsequent iteration to improve the generated image. \fig{fig:refine} illustrated a concrete example for this step.

\subsection{Refinement by Diffusion Re-denoising}
\label{ssec:finetune}

Based on the MLLM feedback, our framework refines the image based on the fact that diffusion models iteratively add details to an image through a reverse diffusion process. Thus, to correct specific parts of an already-generated image, we can introduce controlled noise to the image and rerun the diffusion process with the updated prompt. This approach preserves the integrity of the existing content while selectively refining certain areas.

Recall that a diffusion model is composed of two procedures: forward diffusion and reverse diffusion. In the forward diffusion phase, a clean image $I_0$ is gradually corrupted by adding Gaussian noise across a sequence of time steps $t = 1, 2, \cdots, T$: $I_t = \sqrt{\bar{\alpha}_t} I_0 + \sqrt{1 - \bar{\alpha}_t} \mathcal{N}(0, 1),$
where $\bar{\alpha}_t$ controls the noise level at each time step, and $\mathcal{N}(0, 1)$ represents standard Gaussian noise.

The reverse diffusion process recovers the original image from a Gaussian noise. The model iteratively denoises it based on a noise prediction model $\epsilon(I_t, t)$. By conditioning this noise prediction model also on the text prompts, the model can generate images aligned with specific descriptions.

To incorporate the feedback from the MLLM, we perform a {\it partial re-denoising} process. Instead of regenerating the entire image, we add controlled noise to the existing image such that it matches the noise level of a particular diffusion step $t'$. This noise perturbs the image just enough to allow for modifications while preserving its core structure. Next, we rerun the reverse diffusion process using the refined prompt from MLLM to selectively correct details without losing the overall content. \fig{fig:pipeline} shows an refined image with $t' = T - 2$. We ablate the choice of $t'$ in Appendix~\ref{sec:app:results:redenoising}.

\section{Experiments}
% \subsection{Setup}

We compare \model with state-of-the-art text-to-image generation models on the \data dataset. Additionally, we carry out ablation studies on the incorporation of hierarchical concept decomposition and the number of re-denoising steps.

% gpt-by-topic
\begin{table*}
    \centering\scriptsize
    \setlength{\tabcolsep}{1.4pt} % Adjust column spacing to help fit
    \begin{tabular}{lccccccccccccccc}
    \toprule
         & \multicolumn{5}{c}{{\bf Functional Relation}} & \multicolumn{5}{c}{\bf Multi-subject Interaction} & \multicolumn{5}{c}{\bf Compositional Relation} \\
         \cmidrule(lr){2-6}  \cmidrule(lr){7-11}  \cmidrule(lr){12-16}
         & \text{Human} & Qwen2.5 & ImReward & CLIPS. & B-VQA 
         & \text{Human} & Qwen2.5 & ImReward & CLIPS. & B-VQA 
         & \text{Human} & Qwen2.5 & ImReward & CLIPS. & B-VQA \\
        \midrule
        \textbf{SD}       & 3.360  & 4.480 & 1.657 & 0.971 & 0.366 & 3.225 & 3.800 & 1.171 & 0.894 & 0.306 & 3.583 & 4.867 & 1.415 & 0.898 & 0.379 \\
        \quad \textbf{+ GPT Rewrite} & 3.770  & 4.440 & 1.546 & 0.921 & 0.245 & 3.275 & 3.900 & 1.101 & 0.890 & 0.292 & 3.167 & 4.467 & 1.283 & 0.867 & 0.325 \\
        \quad \textbf{+ GPT Refine}  & 3.450  & 4.600 & 1.524 & 0.951 & 0.290 & 3.175 & 3.600 & 1.022 & 0.858 & 0.292 & 3.667 & 4.800 & 1.390 & 0.889 & 0.389 \\
        \quad \textbf{+ Inf Scale}   & 3.270  & 4.560 & 1.718 & 0.985 & 0.434 & 3.375 & 4.400 & 1.149 & 0.903 & 0.302 & 3.650 & 4.800 & 1.538 & 0.912 & 0.423 \\
        \textbf{DALL·E 3} & 3.940  & 4.680 & 1.535 & 0.880 & 0.226 & 3.775 & \textbf{4.700} & 1.111 & 0.813 & 0.286 & 3.433 & 4.867 & 1.382 & 0.838 & 0.367 \\
        \textbf{DetailScribe} & \textbf{4.280}  & \textbf{4.880} & \textbf{1.761} & \textbf{0.998} & \textbf{0.449} & \textbf{3.800} & 4.600 & \textbf{1.326} & \textbf{0.907} & \textbf{0.343} & \textbf{4.283} & \textbf{5.000} & \textbf{1.545} & \textbf{0.923} & \textbf{0.485} \\
        \bottomrule
    \end{tabular}
    % }
    \vspace{-0.5em}
    \caption{Average human/MLLM likert scale (1 - 5) and pre-trained metrics on three scenarios of sampled \data dataset. We report the human Likert scale (Human Evaluation), MLLM evaluation score (Qwen2.5-VL-32B), as well as ImageReward (ImReward), CLIPScore (CLIPS.) and BLIP-VQA (B-VQA) score. \model receives the highest scores according to human preference in all scenarios.}
    \label{tab:likert}
    \vspace{-1em}
\end{table*}

\subsection{Baselines and Implementation Details}

We compare our model with the following models:

\begin{itemize}
\item \textbf{Stable Diffusion.} We generate the image conditioned on the prompts in \data using the Stable Diffusion SD3.5-large model (\textbf{SD}). We used the SD3.5-large model~\cite{esser2024scalingrectifiedflowtransformers} for all the baselines that are based on a pre-trained T2I generative model.
\item \textbf{Refinement-Augmented Generation} We also include two common strategies for refinement-augmented generation (\textbf{SD + GPT Rewrite} and \textbf{SD + GPT Refine}). For \textbf{SD + GPT Rewrite}, we first use GPT-4o to generate a detailed text prompt given the initial prompt from \data, akin to the concept decomposition used in \model. This improved text prompt is then used to re-generate images with the same pre-trained T2I model. For \textbf{SD + GPT Refine}, we follow the strategy in \cite{ionio2025}. Specifically, GPT-4o was adopted to provide a refined prompt based on an initially generated image and the refinement request prompt which is used in \cite{ionio2025}. We then re-generated the image with the refined prompt.
\item \textbf{Inference scaling ~\citep{ma2025inferencetimescalingdiffusionmodels}.} We implemented a toy version of noise searching in ~\citet{ma2025inferencetimescalingdiffusionmodels} (\textbf{SD + Inf Scale}). Specifically, we sampled two different noises and adopted CLIPScore~\citep{hessel2022clipscorereferencefreeevaluationmetric} as its verifier.
\item \textbf{DALL·E \cite{ramesh2022hierarchicaltextconditionalimagegeneration,bubeck2023sparksartificialgeneralintelligence}} internally integrates LLMs (GPT) to refine prompts with detail model interprets effectively before generating images. We include the DALL·E 3 as a strong baseline to assess \model's advancement in interpreting and generating scenes with rich entity interactions.
\item \textbf{\model.} Our \model implementation leverages Stable Diffusion 3.5 as the foundational model for both image generation and refinement. To ensure fair comparisons, all approaches involving MLLMs and LLMs use separate, identical prompts with the same GPT-4o model, maintaining consistency across evaluations.
\end{itemize}

Among all the algorithms, \textbf{SD + GPT Refine} and inference scaling (\textbf{SD + Inf Scale}), as well as DetailScribe, require two times the computation of the base model (SD) for one iteration of refinement (We neglected the critique time of the MLLM done by commercial APIs, which has a runtime of 10\% of SD3.5.) \textbf{SD + GPT Rewrite} refines prompt unconditional on previous generation, thus requires the same computation as SD.

\subsection{Result}

We evaluate the models on three scenarios from the \data dataset and report the results separately. Due to the scalability of high-quality human evaluation, we sampled 50 prompts from \data for both human evaluation and automatic evaluation (\tbl{tab:likert}), and compared the agreement in between. We adopted Qwen2.5-VL-32B~\citep{bai2025qwen25vltechnicalreport} as our MLLM evaluator to avoid evaluation bias. Overall, MLLM evaluator achieves the highest agreement at 87.6\%, compared to the other metrics: ImageReward (73.6\%), CLIPScore (70.4\%), and BLIP-VQA (67.6\%). We further presented the automatic evaluation on the entire \data in \tbl{tab:scores_entire}. \model outperformed all methods based on SD3.5 in all evaluation.

\begin{table}[!htbp]
    \centering\small
    \setlength{\tabcolsep}{2.8pt}
    % \vspace{-1em}
    % \resizebox{\columnwidth}{!}{
        \begin{tabular}{lcccc}
        \toprule
             & \textbf{Qwen2.5}
             & \textbf{ImReward}
             & \textbf{CLIPS.} 
             & \textbf{B-VQA} \\
             \midrule
                \textbf{SD}                & 4.228 & 1.323 & 0.902 & 0.336 \\
                \textbf{+ GPT Rewrite}     & 4.116 & 1.193 & 0.880 & 0.268 \\
                \textbf{+ GPT Refine}      & 4.141 & 1.255 & 0.880 & 0.300 \\
                \textbf{+ Inf Scale}       & 4.126 & 1.354 & 0.922 & 0.365 \\
                \textbf{DALL·E 3}          & 4.519 & 1.222 & 0.860 & 0.312 \\
                \textbf{DetailScribe}      & \textbf{4.570} & \textbf{1.460} & \textbf{0.923} & \textbf{0.381} \\
                \bottomrule
        \end{tabular}
        % }
    \vspace{-0.5em}
    \caption{Automatic evaluation on entire \data. \model outperforms all baselines on all metrics. We include evaluation by scenario in Appendix D.}
    \label{tab:scores_entire}
    \vspace{-1em}
\end{table}

\fig{fig:benchmark} shows more examples generated by our model and the baselines. As shown in the figure, \model is able to generate images with fine details delineating entity interactions. For example, the first row demonstrates the capability of \model in capturing functional relations. Given the prompt ``A cat sails across the sea in a large seashell, holding a mast.'', all baselines fail to capture the relation ``holding'', while \model is able to generate accurate details of a cat holding the mast with fine details.

The 4th row contains a challenging example of a complex scene layout ``zig-zag path''. \model is the only model capable of generating such fine layout patterns. Stable Diffusion can generate an image with zigzag patterns but fails to reveal a path. Both SD with GPT rewritten and refined prompt and DALL·E fail to follow the prompt on the zigzag pattern and only generate a path with leaves.

\subsection{Ablation: Hierarchical Concept Decomposition Improves Error Detection}
\begin{figure}
   \centering
   \includegraphics[width=1\linewidth]{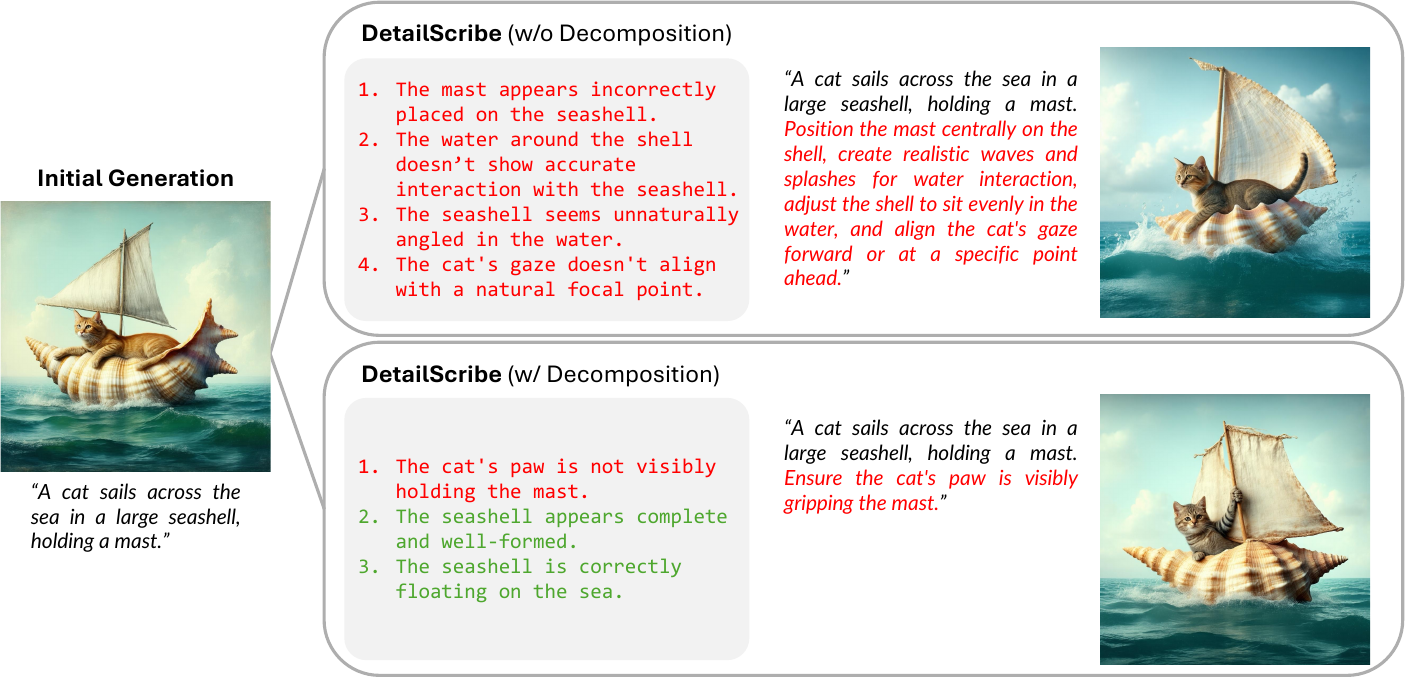}
   \caption{An illustrative example showing the effectiveness of the explicit concept decomposition module. MLLM first critiques the original generation, and identifies the features needs to be correct (red) and the features non-necessary for further modification (green), and then provides the corrected prompt for re-denoising.}
   \vspace{-1em}
   \label{fig:decomposition}
\end{figure}

To evaluate the effectiveness of our hierarchical concept decomposition component, we compare the generated critiques and, subsequently, the refined images with and without our decomposed concepts module. We present the quantitative evaluation on sampled \data in \tbl{tab:decomposition-result}.

\fig{fig:decomposition} shows an illustrative example. The top box in the figure shows critiques generated without our decomposed concepts. The MLLM with access to only the initial prompt does not properly attend to the detailed properties. Instead, the critiques focus more on global attributes such as the shape of objects, lighting conditions, or object arrangements. With the concept decomposition step explicitly added, the MLLM can generate better critiques by attending to local details such as ``missing a spoon'' or action concepts such as ``stirring''. With more errors detected and included in the refined prompt, we also see an improvement in the re-denoising image refinement.

\begin{table}[ht]
    \centering\small
    \setlength{\tabcolsep}{0.8pt}
    % \vspace{-1em}
    % \resizebox{\columnwidth}{!}{
        \begin{tabular}{lccccc}
        \toprule
             & \textbf{Human}
             & \textbf{Qwen2.5}
             & \textbf{ImReward}
             & \textbf{CLIPS.} 
             & \textbf{B-VQA} \\
             \midrule
             \textbf{w/o. Decomp}   & 3.843  & 4.660  & 1.586  & 0.953  & 0.410  \\
             \textbf{w/. Decomp}    & \textbf{4.187}  & \textbf{4.760}  & \textbf{1.609}  & \textbf{0.957}  & \textbf{0.438}  \\
             \bottomrule
        \end{tabular}
        % }
    \vspace{-0.5em}
    \caption{Ablation study on the effectiveness of the concept decomposition module. Explicit concept decomposition significantly improves the generation quality.}
    \label{tab:decomposition-result}
    \vspace{-1em}
\end{table}

\subsection{Ablation: Re-Denoising Step}
\label{sec:app:results:redenoising}

\begin{figure}[ht]
       \centering
       \includegraphics[width=1\linewidth]{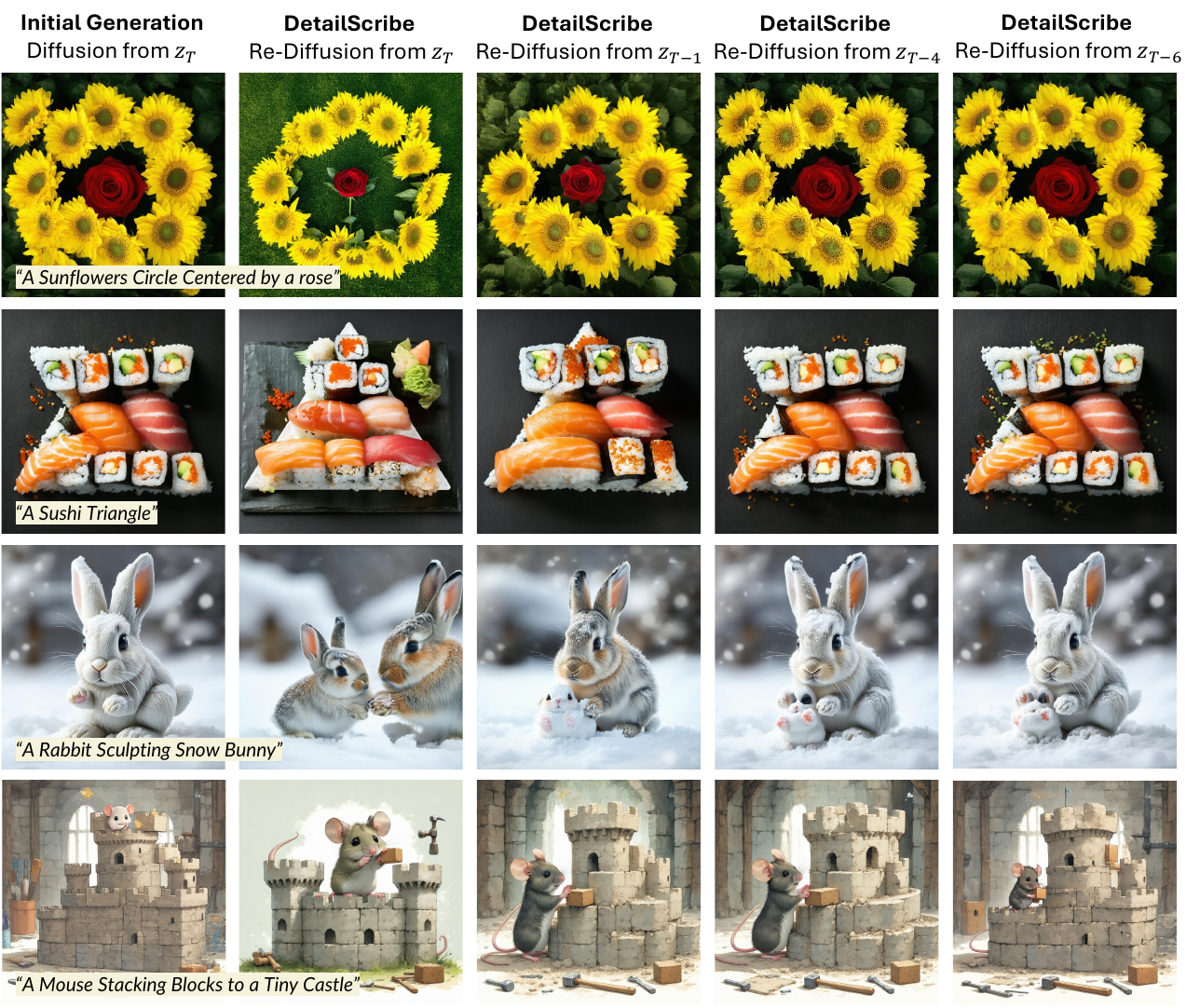}
        \vspace{-1em}
       \caption{Ablation on the number of re-denoising steps. The first column shows the result without the refined prompt. Column 2-5: Starting the re-denoising at step $T$, $T-1$, $T-4$, and $T-6$.}
       \label{fig:rediffusionstep}
        \vspace{-1em}
\end{figure}

We also study the impact of using different numbers of re-denoising steps, as shown in \fig{fig:rediffusionstep}. There are tradeoffs between re-denoising steps and editing level. When $t'$ is set to a very large value (e.g., $T$, corresponding to complete noise), although we can correct the identified errors, the entire image is effectively regenerated. This often introduces new errors. By contrast, if $t'$ is set too low (\ie, when the image is already nearly clean), the updates from the refined prompt have minimal impact, making it difficult to incorporate necessary corrections. To study the tradeoffs, We generate images with SD3.5, starting from steps $T$, $T-1$, $T-4$, and $T-6$, respectively. We ablate the choice of $t'$ in detail in the experiment section. Introducing the refinement prompt at a later de-noising step results in images that are more similar to the original ones, as the diffusion model has fewer steps to refine the generated results. Overall, we find that salient and global attributes of objects, such as the shapes and colors of large objects, are less likely to be modified if the re-denoising occurs at a late stage. However, it is still possible to make local changes that do not interact with large regions of the image, such as adding small objects. In general, for Stable Diffusion models, we found that setting $t'$ close to $T - 2$ gives the best performance. 

We also evaluate the performance of generating images from pure noise using the refined prompt. Given that the refined prompts contain more details, we empirically observe that diffusion models are more prone to missing concepts or concept leakage, such as missing entities (\eg ignoring ``Snow Bunny'' given the prompt ``A Rabbit Sculpting Snow Bunny''). Thus, our refinement-augmented generation procedure can also be interpreted as a coarse-to-fine generation process, making it easier for the model to generate images with coherent global structures and detailed local attributes simultaneously. Furthermore, while we find that a one-round refinement is sufficient, future work may explore extending the framework to multi-round refinements.

\section{Conclusion}

In this paper, we introduce \textit{\data}, a comprehensive dataset focused on fine-grained interactions as a complement to existing text-to-image benchmarks. While most of the previous approaches failed to generate accurate details, we proposed {\it \model}, a generate-then-refine framework that leverages hierarchical critiques from MLLMs to iteratively refine text-to-image generations. We evaluate different algorithms for T2I generation and refinement on our interaction-rich dataset \data, and demonstrate that our model \model achieves superior semantic accuracy and visual coherence.

\xhdr{Limitation.}
To ensure the effectiveness of refinement, we prevent introducing significant changes to the image content by avoiding re-denoising from complete noise. A current limitation of the \model framework, therefore, is its reliance on that the image generated without prompt refinement has a correct global scene structure. For example, if the generated image based on the user input completely misses one of the main subjects. In such cases, even if the MLLM detects errors in the generated image, the re-denoising process may not fix them. Future work may explore seed search~\cite{ma2025inferencetimescalingdiffusionmodels} based on a similar critique strategy.
\section*{Acknowledgements}

We thank Xiaolin Fang, Lei Xu and Lijie Fan for helpful discussion and proofreading. We thank Bowen Pan, Jiejun Jin, Wendi Gao and Yifan Le for volunteering high-quality human evaluation.

\noindent We gratefully acknowledge support from NSF grant 2214177; from AFOSR grant FA9550-22-1-0249; from ONR MURI grant N00014-22-1-2740; and from ARO grant W911NF-23-1-0034; from MIT Quest for Intelligence; from the Boston Dynamics AI Institute; from ONR Science of AI; and from Simons Center for the Social Brain. Any opinions, findings, and conclusions or recommendations expressed in this material are those of the authors and do not necessarily reflect the views of our sponsors.

% \section*{Acknowledgments}

% This document has been adapted
% by Steven Bethard, Ryan Cotterell and Rui Yan
% from the instructions for earlier ACL and NAACL proceedings, including those for
% ACL 2019 by Douwe Kiela and Ivan Vuli\'{c},
% NAACL 2019 by Stephanie Lukin and Alla Roskovskaya,
% ACL 2018 by Shay Cohen, Kevin Gimpel, and Wei Lu,
% NAACL 2018 by Margaret Mitchell and Stephanie Lukin,
% Bib\TeX{} suggestions for (NA)ACL 2017/2018 from Jason Eisner,
% ACL 2017 by Dan Gildea and Min-Yen Kan,
% NAACL 2017 by Margaret Mitchell,
% ACL 2012 by Maggie Li and Michael White,
% ACL 2010 by Jing-Shin Chang and Philipp Koehn,
% ACL 2008 by Johanna D. Moore, Simone Teufel, James Allan, and Sadaoki Furui,
% ACL 2005 by Hwee Tou Ng and Kemal Oflazer,
% ACL 2002 by Eugene Charniak and Dekang Lin,
% and earlier ACL and EACL formats written by several people, including
% John Chen, Henry S. Thompson and Donald Walker.
% Additional elements were taken from the formatting instructions of the \emph{International Joint Conference on Artificial Intelligence} and the \emph{Conference on Computer Vision and Pattern Recognition}.

% % Bibliography entries for the entire Anthology, followed by custom entries
% %\bibliography{anthology,custom}
% % Custom bibliography entries only
\bibliography{custom}

% \appendix

\clearpage
\appendix
\onecolumn

\setcounter{page}{1}

\begin{center}
    \Large \bf Appendix
\end{center}

The appendix is organized as follows. In \sectapp{sec:app:mllm-prompts}, we describe the text prompts for the vision language model (GPT-4o) during the concept decomposition stage and the critique-and-refinement generation stage. In \sectapp{sec:app:data-prompts}, we show the text prompts for interaction topic sampling and completion of \data. In \sectapp{sec:app:evaluation}, we provide details of evaluation guidance, the quantitative automatic evaluation per scenario, and examples for the comparison of different metrics. In \sectapp{sec:app:results}, we provide more qualitative examples generated by \model and additional discussions on the failure modes of the system. Finally, in \sectapp{sec:app:dataset}, we describe the full dataset of our text prompts.

\section{\model Prompting Implementation Details}
\label{sec:app:mllm-prompts}
\subsection{Prompt: Concept Decomposition}

The following is the input prompt to GPT-4o model for the concept decomposition task. Throughout the entire dataset, we consistently use the single example of ``cooking'' as the single-shot example.

\begin{boxA}
We can decompose each abstract concept into interactions defined by contact points and contact objects. \\For example,\\
(concept: cooking)= \\
(hand hold the handle of a ladle) + (ladle stir the ingredient in the pot) + (pot is on a stove)\\
\\
Please do the same for the following concepts in the same format without explanation.\\
Keep the program simple. Use only the most necessary parts of the schema that can be mapped to objects in an image. Describe only the interactions that can happen simultaneously.\\%

\textit{\textbf{[Concepts]}}
\end{boxA}

The following example shows an example input to the large language model.

\begin{boxA}
1.  Topic: polar-bear-cut-cake\\
    Prompt: An anime of a polar bear carefully cutting a berry cake.
\par\quad\par
2.  Topic: Inside a Circle\\
    Prompt: A circle of sunflowers with a single, vibrant red rose in the very center, surrounded by the larger yellow blooms.
\end{boxA}

The following is the response generated by the GPT-4 model based on the previous inputs.

\begin{boxA}
1.  Topic: polar-bear-cut-cake\\
    Prompt: An anime of a polar bear carefully cutting a berry cake.\\
    Components: (concept: polar-bear-cut-cake) = (paw hold knife) + (knife cut through cake) + (cake rest on plate)
    \par\quad\par
2.  Topic: Inside a Circle\\
    Prompt: A circle of sunflowers with a single, vibrant red rose in the very center, surrounded by the larger yellow blooms.\\
    Components: (concept: circle) = (sunflowers form circle) + (red rose placed in center) + (yellow blooms surround rose)", "description": "A circle of sunflowers with a single, vibrant red rose in the very center, surrounded by the larger yellow blooms.
\end{boxA}

\subsection{Prompt: Critique and Prompt Refinement}

The following is the input prompt to the GPT-4o model for the critique and prompt refinement task. In this stage, we do not provide any in-context learning example to the model, other than instructions and the concept decomposition generated in the previous step.

\begin{boxA}
This is an image generated with the prompt: {\it \textbf{[prompt]}}. But this image looks bizarre. Examine the image carefully follow the concept of \textit{ \textbf{[topic]}} attached below and other components in the image. For each abnormal part, describe what is wrong with it, then give a concise description on how to correct it. \\\\
Components: {\itshape \textbf{[Components]}}\\\\
Do not simply rely on the components described above, but also exam whether an object looks complete.\\
First, write your answer in a numbered list,\\
Then, rank the issues by their degree of impact on presenting the concept.\\
Last, summarize the correction instructions in order, and write a new description with the first sentence to be {\itshape \textbf{[prompt]}} followed by correction instructions.\\\\
Do not change the first sentence.\\
Be concise, no more than 70 words, but make sure not to miss any information that needs to be corrected.\\
Provide the new description in angle brackets \textless\textgreater.\\
The components described in the original prompt are essential, do not question the concepts in the original prompt.
\end{boxA}

\section{\data Generation Details}
\label{sec:app:data-prompts}

We adopted GPT-4o to automatically generate prompt in InterActing dataset. We first prompt the LLM to generate a list of topics for each scenario by providing examples for in-context learning. Then, we call the API to complete the prompt in \data one by one.
\subsection{Functional and Action-Based Interactions}
\subsubsection{Topic Generation: Tool Manipulation}
\begin{boxA}
    Given a tool manipulation action, we can create some novel and previously unseen scene or cartoon that can be present by an image. For example,\\
    \\
    Concept: Cut-Cake\\
    Tool: knife\\
    Image description: n anime of a polar bear carefully cutting a berry cake.\\
    \\
    Think of concepts similar to cut-cake, carve-wood, cut-pizza, paint-portrait. Provide 150 different but similar concepts, separate them by comma ','.\\
    All lowercase please.
\end{boxA}
\subsubsection{Topic Generation: Physical Contact}
\begin{boxA}
    Given an action has direct physical contact, we can create some novel and previously unseen scene or cartoon that can be present by an image. For example,\\
    \\
    Concept: sculpting-snow\\
    Image description: A rabbit carefully sculpts a tiny snow bunny with its paws, adding details like ears and whiskers to the figure.\\
    \\
    Think of concepts similar to stacking, holding. Provide 150 different but similar concepts, separate them by comma ','.\\
    All lowercase please.
\end{boxA}
\subsubsection{Prompt Completion}
\begin{boxA}
    Come up with a description of an animal \textbf{\textit{[content]}}, the description should be similar as the following example and uncommon to be observed. Do not use passive voice. \\
    Double check the description to focus on major relation, which is \textbf{\textit{[content]}}.
    Write down your answer in this format: \{"topic": \textbf{\textit{[content]}}, "prompt": description\} \\
    For example: \\
    interaction: "taking photos" \\
    Entities: squirrel \\
    Description: A squirrel taking photos with a camera.\\
    Then, the output should be: \{"topic": "taking-photos", "prompt": "A squirrel taking photos with a camera."\}
\end{boxA}

\subsection{Multi-subject Interactions}
\subsubsection{Topic Generation}
\begin{boxA}
    Given a verb of 2 subjects' interaction, we can create some novel and previously unseen scene or cartoon that can be present by an image. For example,\\
    \\
    Concept: High-Fiving\\
    Image description: A dolphin and a seal leap from the water, high-fiving with their flippers.
    \\
    Think of concepts similar to High-Fiving, Lifting-Togethe, huddling-for Warmth. Provide 100 different but similar concepts, separate them by comma ','.\\
    All lowercase please.
\end{boxA}

\subsubsection{Prompt Completion}
\begin{boxA}
    Come up with a description of two animals doing \textbf{\textit{[content]}}, the description should be similar as the following example and uncommon to be observed. Do not use passive voice.\\
    Double check the description to focus on major relation, which is \textbf{\textit{[content]}}.
    Write down your answer in this format: \{"topic": \textbf{\textit{[content]}}, "prompt": description\}\\
    The description must contains the exact \textbf{\textit{[content]}} word.\\
    For example:\\
    Concept: "High-Fiving"\\
    Description: A dolphin and a seal leap from the water, high-fiving with their flippers.\\
    Then, the output should be: \{"topic": "High-Fiving", "prompt": "A dolphin and a seal leap from the water, high-fiving with their flippers."\}
                    
\end{boxA}

\subsection{Compositional Spatial Relationships}
\subsubsection{Topic Generation}
We generated the abstract layouts and geometric patterns together and use classify them manually with the assistance of LLM.
\begin{boxA}
    Given an abstract concept, we can create some novel scene that can be present by an image. For example,\\
    \\
    Concept: tic-tac-toe\\
    Image description: A tic-tac-toe composed by tomato and cucumber as the players symbols.\\
    \\
    Think of concepts similar to tic-tac-toe, atom, triangle, tree. Provide 300 different but similar concepts, separate them by comma ','.\\
    All lowercase please.
\end{boxA}
\subsubsection{Prompt Completion}
\begin{boxA}
    Come up with a description of a scene which is a novel combination of \textbf{\textit{[content]}}, the description should be similar as the following example and uncommon to be observed. Do not use passive voice. \\
    Double check the description to focus on major relation, which is \textbf{\textit{[content]}}.
    Write down your answer in this format: \{"topic": \textbf{\textit{[content]}}, "prompt": description\} \\
    For example: \\
    Concept: "tic-tac-toe" \\
    Description: A tic-tac-toe composed by tomato and cucumber as the players symbols. \\
    Then, the output should be: \{"topic": "tic-tac-toe", "prompt": "A tic-tac-toe composed by tomato and cucumber as the players symbols."\}
\end{boxA}

\section{Model Evaluation Details}
\label{sec:app:evaluation}

In this section, we include the evaluation guidelines for human and MLLM annotators, and then present automatic evaluation by scenario based on the MLLM rating, ImageReward~\cite{xu2023imagerewardlearningevaluatinghuman}, CLIPScore~\cite{hessel2022clipscorereferencefreeevaluationmetric} and BLIP-VQA~\cite{huang2023t2icompbenchcomprehensivebenchmarkopenworld} (\tbl{tab:auto_eval_entire_by_scenario}). We also provide running examples for comparison of human evaluation and automatic evaluation.

\subsection{Human evaluation}
% \xinyi{move to sup. if space limit}
In the human evaluation process, we asked annotators to rate images on a Likert scale. For each prompt, annotators were presented with images generated by all models, which were randomly shuffled to mitigate order bias. Annotators were instructed to assign a score from 1 to 5 based on image-text alignment, following these guidelines:
\begin{itemize}
    \item 1: The image is completely irrelevant to the prompt.
    \item 2: The image contains some relevant objects, but they exhibit severe issues (e.g., distortion or missing parts).
    \item 3: The image includes most relevant objects, but some elements implied by the prompt are missing (e.g., missing critical tools or patterns to complete the interaction).
    \item 4: The image captures most relevant objects and infers additional ones successfully, but there are minor issues with object relationships (e.g. improvement to appearance of tools, subpart, or limbs needed.)
    \item 5: The image accurately and naturally reflects the prompt description.
\end{itemize}

To reduce variance and the impact of instruction ambiguity, we recruited four volunteers for this experiment. For each annotator, we calculated the average score for each model.

\subsection{MLLM Evaluation Prompt Implementation Details}
We prompt the MLLM to assign a score from 1 to 5 based on the following guidelines.
\subsubsection{Functional and Action-Based Interactions, Multi-subject Interactions}
\begin{boxA}
The above images were generated with the prompt: {\it \textbf{[prompt]}}.\\
Please rate the text-image alignment score of each image from 1 to 5, focusing on {\it \textbf{[topic]}} and follow the criteria: \\\\
1: poor interaction, subject(s) not acting correctly.\\
2: subject(s) incorrect/inaccurate.\\
3: critical part missing (e.g., missing critical tools or patterns to complete {\it \textbf{[topic]}}).\\
4: nearly perfect but some subparts need further improvement (e.g., needs to refine appearance of tools or limbs, subject is not {\it \textbf{[topic]}} correctly).\\
5: image perfectly depicts {\it \textbf{[topic]}}.\\\\
Return the score in angle brackets \textless\textgreater. For example, if the image is nearly perfect and got score 4, response: \textless4\textgreater
\end{boxA}

\subsubsection{Compositional Spatial Relationships}
\begin{boxA}
You are my assistant to identify objects and their spatial layout in the image. According to the image, evaluate if the text {\it \textbf{[prompt]}} is correctly portrayed in the image.\\

Give a score from 1 to 5 according to the criteria: \\
5: correct spatial layout ({\it \textbf{[topic]}}) in the image for all objects mentioned in the text.\\
4: basically, spatial layout of objects matches the text.\\
3: spatial layout not aligned properly with the text.\\
2: image not aligned properly with the text.\\
1: image almost irrelevant to the text.\\

Return the score in angle brackets \textless\textgreater. For example, if the image's spatial layout of objects matches the text and got score 4, response: \textless4\textgreater
\end{boxA}

\subsubsection{Automatic evaluation Results}
\begin{table*}[ht]
    \centering\small
    \setlength{\tabcolsep}{2pt} % Adjust column spacing to help fit
    \resizebox{\textwidth}{!}{ % Resize to fit the full width
    \begin{tabular}{lcccc cccc cccc}
        \toprule
         & \multicolumn{4}{c}{\textbf{Functional Relation}} 
         & \multicolumn{4}{c}{\bf Multi-subject Interaction} 
         & \multicolumn{4}{c}{\bf Compositional Relation} \\
         \cmidrule(lr){2-5}  \cmidrule(lr){6-9}  \cmidrule(lr){10-13}
         & Qwen2.5 & ImReward & CLIPS. & B-VQA 
         & Qwen2.5 & ImReward & CLIPS. & B-VQA 
         & Qwen2.5 & ImReward & CLIPS. & B-VQA \\
        \midrule
        \textbf{SD}                & 4.271 & 1.471 & 0.914 & 0.430 & 3.780 & 1.247 & 0.917 & 0.203 & 4.550 & 0.949 & 0.851 & 0.184 \\
        \quad \textbf{+ GPT Rewrite}  & 4.133 & 1.285 & 0.881 & 0.323 & 3.735 & 1.206 & 0.910 & 0.198 & 4.445 & 0.902 & 0.846 & 0.171 \\
        \quad \textbf{+ GPT Refine}   & 4.198 & 1.401 & 0.889 & 0.376 & 3.755 & 1.216 & 0.904 & 0.204 & 4.355 & 0.852 & 0.830 & 0.169 \\
        \quad \textbf{+ Multi-seed}   & 4.282 & 1.502 & 0.935 & 0.472 & 3.845 & 1.242 & 0.934 & 0.210 & 4.565 & 1.019 & 0.872 & 0.197 \\
        \textbf{DALL·E 3}            & 4.477 & 1.343 & 0.869 & 0.392 & \textbf{4.310} & 1.139 & 0.868 & 0.193 & \textbf{4.850} & 0.938 & 0.824 & 0.190 \\
        \textbf{DetailScribe}        & \textbf{4.582} & \textbf{1.598} & \textbf{0.936} & \textbf{0.482} & 4.275 & \textbf{1.378} & \textbf{0.935} & \textbf{0.242} & 4.830 & \textbf{1.123} & \textbf{0.875} & \textbf{0.216} \\
\bottomrule
    \end{tabular}
    }
    \vspace{-0.5em}
    \caption{Auto evaluation on entire \data dataset by scenario.}
    \label{tab:auto_eval_entire_by_scenario}
    % \vspace{-1em}
\end{table*}

\subsection{Agreement between human evaluation and automatic Evaluation}

\begin{figure*}[ht]
   \centering
   \includegraphics[width=.9\linewidth]{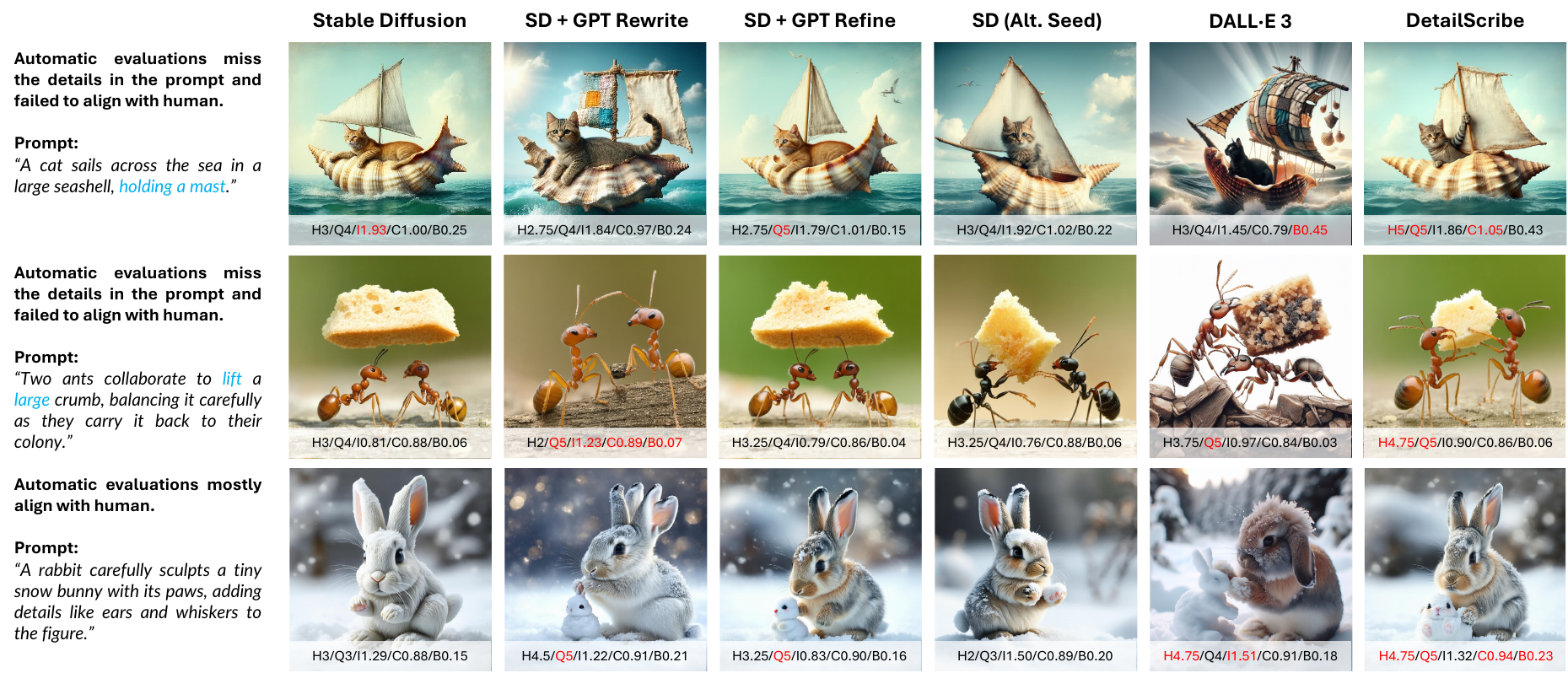}
   \caption{Example of automatic evaluation misinterpreted and successfully interpreted the critical interaction in prompts. The details in the prompt missed by the evaluators are highlighted in blue with an underline. (H: human, Q: Qwen2.5, I: ImageReward, C: CLIPScore, B: BLIP-VQA, Highest score of each metrics are highlighted in red)}
   \label{fig:gpt_scores}
\end{figure*}

We have used the Qwen2.5-VL-32B model as an MLLM-based automatic evaluation metric for comparing different models. The Qwen2.5 model takes the instruction as shown in Appendix C.2.2 and directly outputs a score for the individual images. We further adopted ImageReward ~\cite{xu2023imagerewardlearningevaluatinghuman}, CLIPScore~\cite{hessel2022clipscorereferencefreeevaluationmetric} and BLIP-VQA~\cite{huang2023t2icompbenchcomprehensivebenchmarkopenworld} as our automatic evaluator. In this section, we present detailed examples of automatic judgments on generated images from different models to illustrate their capabilities and limitations in evaluating complex concepts involving multi-entity interactions.
Specifically, in \fig{fig:gpt_scores} we include two example where the automatic evaluator failed to recognize the incorrect intersections between objects or missing key components in the prompt, leading to uniformly high scores for all generated images. Additionally, we provide another example where most evaluators gave consistent ratings. \\
Capable to align with human in most evaluations, MLLM-based evaluator tends to give high scores. These examples highlight the current challenges in using MLLM as a judge for complex compositional reasoning tasks. We recommend human evaluation for future experiment if cost and throughput allows.

% \clearpage

\clearpage
\section{Additional Results and Analysis}
\label{sec:app:results}

In this section, we provide additional qualitative examples generated by different models based on the text instructions from \data. Furthermore, we provide examples and discussions about the effectiveness of concept decomposition and progressive refinement. We also discuss the limitation of the current system in making global scene edits.

\subsection{Qualitative Examples}

\begin{figure}[H]
   \centering
   \includegraphics[width=0.7\linewidth]{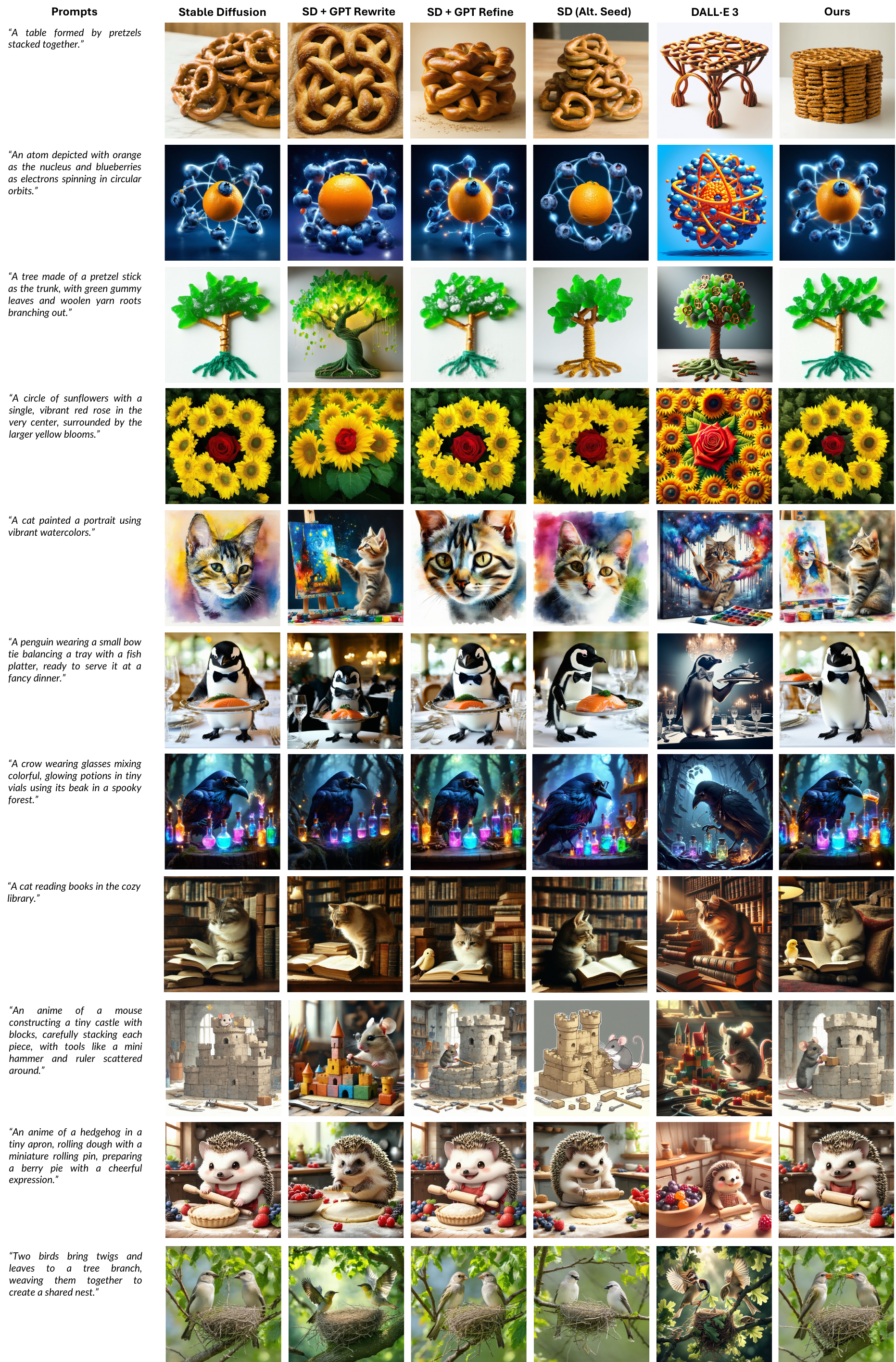}
   \caption{More examples generated by \model and baselines on the \data dataset.}
   \label{fig:more_examples}
   \vspace{-1em}
\end{figure}

\fig{fig:more_examples} provides additional qualitative examples generated by \model and other baselines. Overall, \model is capable of generating faithful images according to different complex language descriptions.

\subsection{Qualitative Studies on the Effect of Concept Decomposition and Progressive Refinement}

\begin{figure}[H]
   \centering
   \rotatebox{90}{ % Rotate by 90 degrees
   \begin{minipage}{.85\textheight}
       \includegraphics[width=.85\textheight]{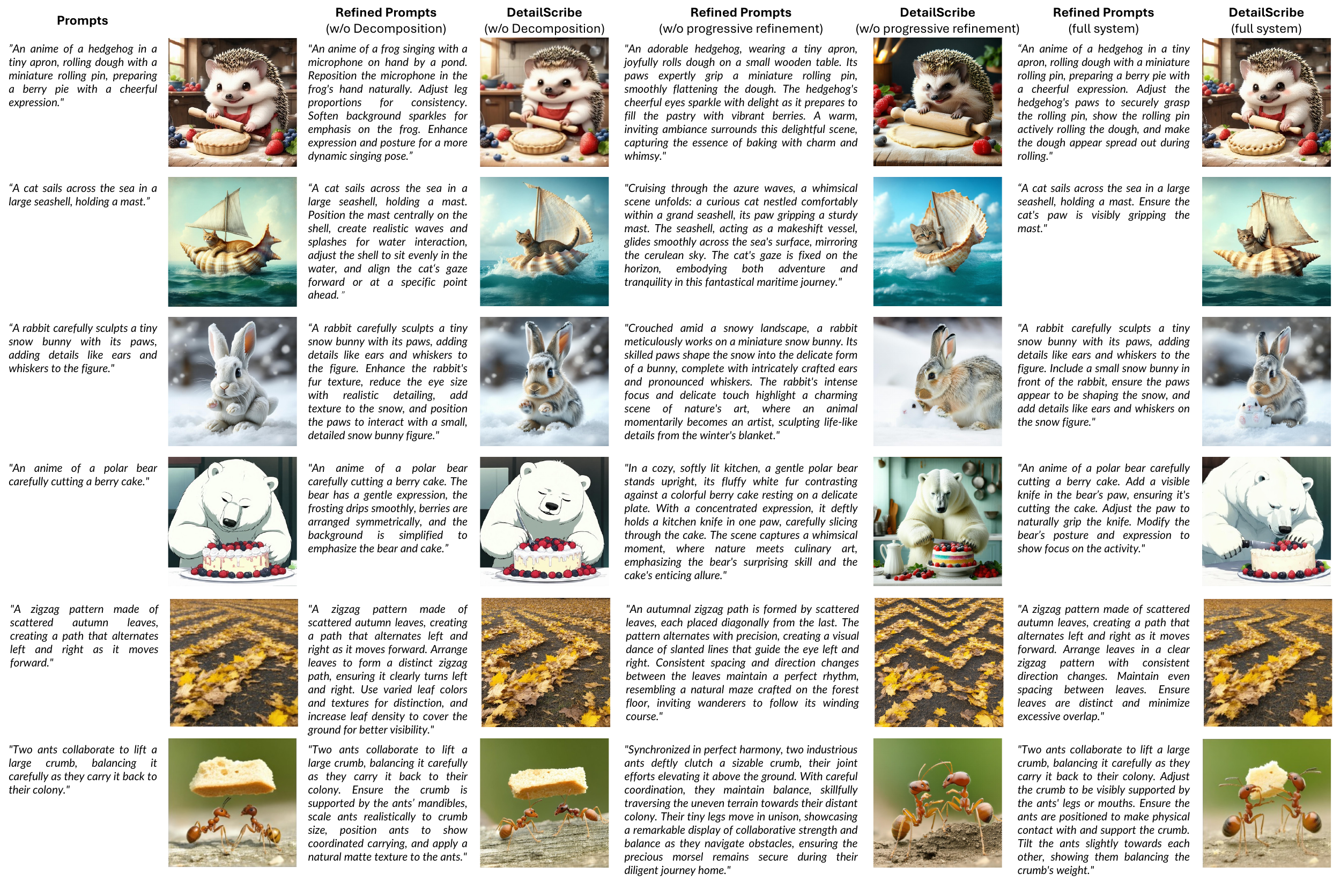}
       \captionsetup{width=0.85\textheight}
       \caption{Qualitative examples that illustrate the effectiveness of concept decomposition and progressive refinement. We compare our full pipeline (shown as full system in the last two columns with two alternatives: one without concept decomposition (shown as w/o Decomposition), and another one without the generate-then-refine procedure (shown as w/o progressive refinement).}
       % \jiayuan{change to w/o progressive refinement}}
       \label{fig:ablation_examples}
       \end{minipage}
    }
\end{figure}

We present a comparative analysis of MLLM-generated critiques and refinements with and without the explicit concept decomposition step. \fig{fig:ablation_examples} illustrates the result. The results indicate that incorporating concept decomposition significantly enhances the MLLM’s ability to focus on meaningful entity interactions rather than overemphasizing fine-grained image details, such as minor variations in subject expressions.

Additionally, we compare our model with a variant approach without progressive feedback, where the image is generated directly from a refined text prompt derived from the input instruction and the concept decomposition. Our pipeline, which integrates both concept decomposition and progressive refinement, consistently outperforms both alternatives by a substantial margin. This highlights the critical role of concept decomposition in structuring the model’s understanding and the refinement procedure in producing faithful and realistic images.

% \clearpage

\subsection{Qualitative Studies on the Effect of Global Seeds}

\begin{figure*}[ht]
   \centering
   \includegraphics[width=1\linewidth]{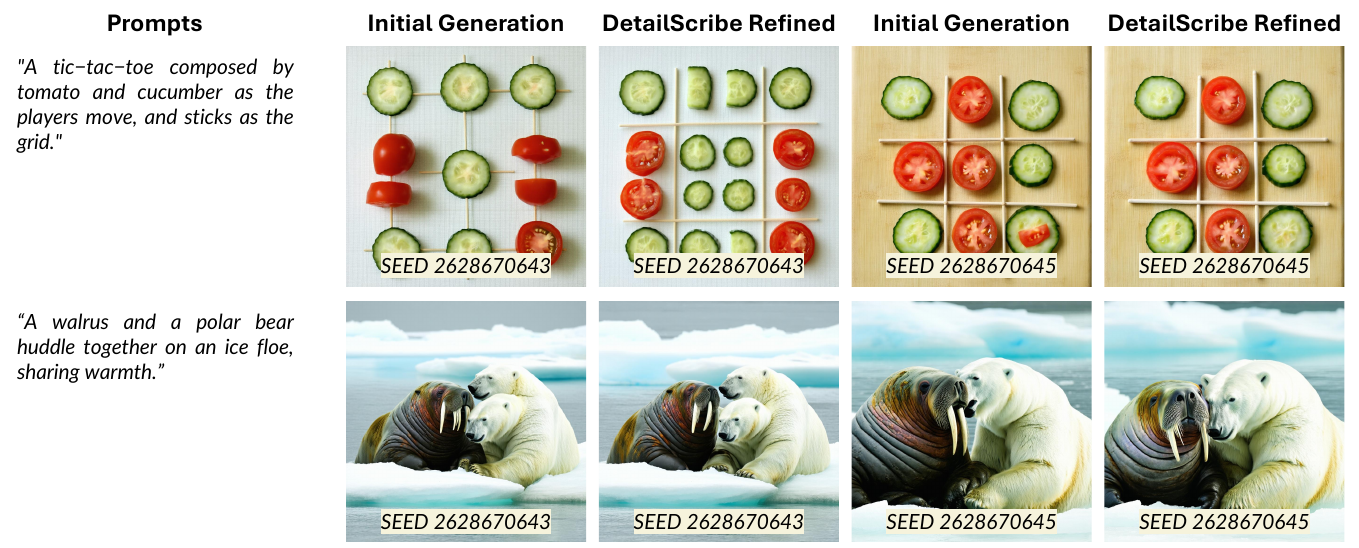}
   \caption{Different global random seeds yield different initial images. They also result in different refined images using the \model framework.}
   \label{fig:alt_seed}
\end{figure*}

A current limitation of the \model framework lies in its reliance on the assumption that the initial image generated without prompt refinement has a roughly correct global scene structure. This limitation arises from the inability of the re-denoising step to introduce large-scale changes to the image, such as adding missing subjects or significantly altering the scene layout. Our examples in \fig{fig:alt_seed} illustrate that, due to the stochasticity inherent in text-to-image generative models, different global random seeds at inference yield different initial images, which in turn affect the final output of our system.
While our approach effectively resolves issues like entity interactions in the image, it struggles with global-scale edits, such as adjusting the global layout or zoom level of the scene. As a potential direction for future work, one can consider sampling multiple initial images simultaneously and performing a post-hoc selection process to identify the most suitable candidates for refinement, leveraging MLLM feedback to improve global scene accuracy.

\clearpage

\section{\data Dataset}
\label{sec:app:dataset}

\subsection{Functional and Action-Based Interactions}
\subsubsection{Statistic of topic}
\begin{itemize}
    \item \textbf{Tool manipulation:}
    octopus paint canvas(1), cat sail(1), fox stir stew(1), squirrel take photo(1), play chess(1), carve wood(1), fly a kite(1), make a potion(1), polar bear cut cake(1), raccoon sweep floor(1), rabbit paint mural(1), bear mop floor(1), painted a portrait(1), reading a book(2), opening a door(1), skiing down(2), slicing(2), skateboarding(2), typing on(2), tying(2), cleaning(1), cooking(2), chopping vegetables(1), dragging(1), skating on(1), photographing(1), baking a cake(1), brushing(1), pouring(1), playing piano(1), playing video games(1), painting(1), writing a book(1), cutting(2), driving(2), using tools(1), constructing bridges(1), playing the piano(1), assemble chair(1), burn wood(1), clip cloth(1), crack coconut(1), decorate cake(1), dip pen ink(1), fold paper crane(1), hang picture(1), measure flour(1), mix dough(1), mount painting(1), paint mural(1), pour syrup(1), saw ice(1), sew button(1), sew patch(1), shape clay(1), shovel snow(1), sketch blueprint(1), sort silverware(1), spray paint(1), spread butter(1), spread paint(1), chop vegetables(1), sharpen pencil(1), drill hole(1), slice bread(1), hammer nail(1), saw plank(1), stir soup(1), sew fabric(1), knit scarf(1), whisk eggs(1), grate cheese(1), peel apple(1), crack walnut(1), roll dough(1), frost cupcake(1), flip pancake(1), grill steak(1), dice carrots(1), scramble eggs(1), boil pasta(1), roast marshmallow(1), squeeze lemon(1), grind coffee(1), sculpt ice(1), weld metal(1), etch glass(1), sand wood(1), string beads(1), carve pumpkin(1), mend socks(1), tie knot(1), cut origami(1), bend wire(1), engrave stone(1), pour tea(1), ladle soup(1), weave basket(1), thread needle(1), sketch bird(1), chisel marble(1), polish shoe(1), blow glass(1), stamp seal(1), dip paintbrush(1), spray graffiti(1), lace shoes(1), brush hair(1), write calligraphy(1), erase mistake(1), clamp wood(1), file nail(1), clip coupon(1), punch hole(1), staple paper(1), tape box(1), tie bow(1), glaze pottery(1), stack books(1), roll sushi(1), fill bottle(1), slice melon(1), scoop ice cream(1), carve sign(1), build clock(1), stretch canvas(1), pluck strings(1), shred paper(1), scrape icing(1), light candle(1), zip jacket(1), unlock door(1), wrap gift(1), rinse brush(1), fold laundry(1), pour candle wax(1), whittle stick(1), cut hair(1), plaster wall(1), glue model(1), mix paint(1), stencil design(1), thread loom(1), cut paper snowflake(1), iron shirt(1), chainmail weaving(1), stamp pattern(1), embroider flower(1), stack firewood(1), filter coffee(1), frost window(1), tie fishing line(1), shape snowball(1), drill teeth(1), paint fence(1), sharpen knife(1), pour molten metal(1), cut ribbon(1), stretch cloth(1), fold napkin(1), heat metal(1), frame picture(1), string violin(1), curl hair(1), carve soap(1), squeeze clay(1), lay tiles(1), weave rug(1), nail shoe(1), stitch wound(1), break egg(1), spread jam(1), poke hole(1), patch clothes(1), snap twig(1), snap chopsticks(1), pick lock(1), snip hedge(1), clean brush(1), file metal(1), press flower(1), scoop sand(1), mold snowman(1), snap photo(1), knot rope(1), chop logs(1), chain carvings(1), stir sauce(1), cut vinyl(1), fold fan(1), tape frame(1), trim trees(1), sift flour(1), stir coffee(1), whisk cream(1), shell peanut(1), pry lid(1), knead bread(1), scrub floor(1), filter water(1), stain wood(1), melt chocolate(1), stir tea(1), light match(1), brush coat(1), tie lace(1), saw bamboo(1), peel corn(1), scrape wax(1), rotate key(1), dip spoon(1), stack bricks(1), weave tapestry(1), bind book(1), decorate mask(1), skim cream(1), pour wine(1), paint sculpture(1) 

    \item \textbf{Physical Contact:}
    gripping(1), clutching(1), hugging(1), clasping(1), twisting(3), lifting(1), balancing(1), squeezing(2), pressing(2), pinching(1), pulling(3), pushing(5), wrapping(2), cupping(1), stroking(1), rubbing(1), kneading(1), twirling(1), braiding(1), weaving(1), shaping(1), molding(1), rolling(1), spinning(2), tapping(2), scratching(1), clapping(1), smudging(1), flicking(2), catching(2), tossing(3), kicking(1), propping(1), supporting(1), cradling(1), wiping(1), polishing(2), dusting(1), scrubbing(1), slapping(1), punching(1), drumming(2), doodling(1), carving(1), shuffling(1), stacking(1), arranging(1), aligning(1), linking(1), snapping(2), unrolling(1), folding(1), creasing(1), peeling(1), popping(1), tying(1), knotting(1), stretching(1), skipping(1), waving(1), shaking(2), fanning(1), poking(1), nudging(1), flipping(1), scooping(1), ladling(1), swiping(2), tugging(2), shoveling(1), sifting(1), spreading(1), smoothing(1), plucking(1), patting(1), scraping(1), slathering(1), dipping(1), drizzling(1), stamping(1), tracing(1), sketching(1), threading(1), embroidering(1), ruffling(1), petting(1), nuzzling(1), tickling(1), adjusting(1), placing(1), tucking(1), clicking(1), rotating(1), juggling(1), whittling(1), cranking(1), filing(1), plating(1), tacking(1), dabbing(1), buffing(1), dancing(2), pouring(1), jostling(1), stirring(1), rabbit sculpt snow(1), heron fishing(1), build(1), bake(1), serve(1), cook(1), frog sing(1), beaver drink(1), rabbit set table(1), squirrel carve acorn(1), fox pour tea(1), cricket write music leaves(1), beaver cut pizza(1), draped over(2), playing guitar(7), playing chess(4), balancing on(2), posing with(2), reflecting in(2), eating at(2), walking up(2), sewn on(2), getting on(1), approaching(2), walking towards(2), walking to(1), growing by(2), grabbing(2), playing music(1), scattered on(1), jumping on(1), climbing(2), pointing at(2), coming down(2), preparing(2), going into(2), decorating(2), growing from(1), washing(2), herding(2), chewing(2), working in(2), picking up(2), looking over(2), shining through(2), smelling(1), running through(1), enclosing(1), going through(1), walking into(1), falling off(1), decorated with(1), walking past(1), towing(1), blowing out(1), jumping off(1), moving(1), running across(1), hang from(1), sitting around(1), cooked in(1), buying(1), standing around(1), growing behind(1), exiting(1), jumping over(1), looking down at(1), looking into(1), leaning over(1), growing next to(1), observing(1), traveling on(1), wading in(1), growing along(1), opening(1), eating in(1), standing against(1), trying to catch(1), stacking rocks(1), lying next to(1), guiding(1), smoking(1), conducting interviews(1), wearing(2), holding(2), sitting on(2), standing on(2), riding(2), standing in(2), lying on(2), hanging on(2), eating(2), looking at(2), covering(1), sitting in(2), hanging from(2), parked on(2), riding on(2), covered in snow(1), flying in(2), sitting at(2), playing with(2), reading(2), reading books(2), filled with laughter(1), crossing(1), swinging(2), standing next to(2), touching(1), flying(2), contain(2), hitting(2), lying in(2), standing by(2), driving on(2), throwing(2), sitting on top of(2), walking down(2), parked in(2), standing near(2), performing tricks(1), printed on(1), facing(2), leaning against(2), grazing on(2), standing in front of(2), drinking(2), topped with(2), swimming in(2), driving down(2), hanging over(2), feeding(2), waiting for(1), running on(2), talking to(1), holding onto(1), eating from(1), perched on(1), parked by(1), hanging above(1), floating on(1), wrapped around(1), near(1), carrying(1), walking on(1), covered in leaves(1), watching(1), covered in(1), enthusiasm(1), ambition(1), walking in(1), surrounded by(1), pulled by(1), growing on(1), standing behind(1), playing(1), mounted on(1), surfing(1), talking on(1), worn on(1), resting on(1), floating in(1), lying on top of(1), playing in(1), walking with(1), pushed by(1), playing on(1), sitting next to(1)

\end{itemize}
\subsubsection{Selected prompts}
\begin{adjustwidth}{2em}{0em}
\begin{enumerate}
    \item \textbf{Topic}: Octopus-Paint-Canvas
    \begin{description}
        \item[\textbf{Prompt}:] An octopus in an art studio is painting on a canvas.
    \end{description}
    \item \textbf{Topic}: Cat-Sail
    \begin{description}
        \item[\textbf{Prompt}:] A cat sails across the sea in a large seashell, holding a mast.
    \end{description}
    \item \textbf{Topic}: Fox-Stir-Stew
    \begin{description}
        \item[\textbf{Prompt}:] A fox stirs a stew in a hollowed-out tree trunk.
    \end{description}
    \item \textbf{Topic}: Squirrel-Take-Photo
    \begin{description}
        \item[\textbf{Prompt}:] A squirrel taking photos with a camera.
    \end{description}
    \item \textbf{Topic}: Rabbit-Sculpt-Snow
    \begin{description}
        \item[\textbf{Prompt}:] A rabbit carefully sculpts a tiny snow bunny with its paws, adding details like ears and whiskers to the figure.
    \end{description}
    \item \textbf{Topic}: Heron-Fishing
    \begin{description}
        \item[\textbf{Prompt}:] A heron fishing by the river.
    \end{description}
    \item \textbf{Topic}: Build
    \begin{description}
        \item[\textbf{Prompt}:] An anime of a mouse constructing a tiny castle with blocks, carefully stacking each piece, with tools like a mini hammer and ruler scattered around.
    \end{description}
    \item \textbf{Topic}: Bake
    \begin{description}
        \item[\textbf{Prompt}:] An anime of a hedgehog in a tiny apron, rolling dough with a miniature rolling pin, preparing a berry pie with a cheerful expression.
    \end{description}
    \item \textbf{Topic}: Play-Chess
    \begin{description}
        \item[\textbf{Prompt}:] An anime of a raven perched on a table, moving pieces on a tiny chessboard with its beak, calculating each move as it faces off against another bird.
    \end{description}
    \item \textbf{Topic}: Carve-Wood
    \begin{description}
        \item[\textbf{Prompt}:] An anime of a beaver wearing a small hat, using its teeth to carve an intricate wooden statue, with wood shavings scattered around.
    \end{description}
    \item \textbf{Topic}: Serve
    \begin{description}
        \item[\textbf{Prompt}:] A penguin wearing a small bow tie balancing a tray with a fish platter, ready to serve it at a fancy dinner.
    \end{description}
    \item \textbf{Topic}: Cook
    \begin{description}
        \item[\textbf{Prompt}:] A bear in a tiny chef hat flipping pancakes in a pan, with jars of honey around, preparing breakfast in the forest.
    \end{description}
    \item \textbf{Topic}: Fly-A-Kite
    \begin{description}
        \item[\textbf{Prompt}:] An elephant holding a vine tied to a leaf-shaped kite, flying it in the air on a breezy day.
    \end{description}
    \item \textbf{Topic}: Make-A-Potion
    \begin{description}
        \item[\textbf{Prompt}:] A crow wearing glasses mixing colorful, glowing potions in tiny vials using its beak in a spooky forest.
    \end{description}
    \item \textbf{Topic}: Polar-Bear-Cut-Cake
    \begin{description}
        \item[\textbf{Prompt}:] An anime of a polar bear carefully cutting a berry cake.
    \end{description}
    \item \textbf{Topic}: Raccoon-Sweep-Floor
    \begin{description}
        \item[\textbf{Prompt}:] An anime of a raccoon sweeping floor with a broom.
    \end{description}
    \item \textbf{Topic}: Frog-Sing
    \begin{description}
        \item[\textbf{Prompt}:] An anime of a frog singing with a microphone on hand by a pond.
    \end{description}
    \item \textbf{Topic}: Beaver-Drink
    \begin{description}
        \item[\textbf{Prompt}:] An anime of a beaver sipping water from a pond through a hollow stick like a straw.
    \end{description}
    \item \textbf{Topic}: Rabbit-Set-Table
    \begin{description}
        \item[\textbf{Prompt}:] An anime of a rabbit setting plates on a rock 'table.'
    \end{description}
    \item \textbf{Topic}: Squirrel-Carve-Acorn
    \begin{description}
        \item[\textbf{Prompt}:] An anime of a squirrel carving a design on an acorn using a chisel.
    \end{description}
    \item \textbf{Topic}: Fox-Pour-Tea
    \begin{description}
        \item[\textbf{Prompt}:] An anime of a fox pouring tea from a tiny pot into cups.
    \end{description}
    \item \textbf{Topic}: Cricket-Write-Music-Leaves
    \begin{description}
        \item[\textbf{Prompt}:] An anime of a cricket scratching musical notes onto a large leaf with a pen.
    \end{description}
    \item \textbf{Topic}: Beaver-Cut-Pizza
    \begin{description}
        \item[\textbf{Prompt}:] An anime of a beaver cutting a pizza.
    \end{description}
    \item \textbf{Topic}: Rabbit-Paint-Mural
    \begin{description}
        \item[\textbf{Prompt}:] An anime of a rabbit painting a colorful mural on a wall.
    \end{description}
    \item \textbf{Topic}: Bear-Mop-Floor
    \begin{description}
        \item[\textbf{Prompt}:] An anime of a bear cleaning its cave floor with a bundle of grass.
    \end{description}
\end{enumerate}
\end{adjustwidth}

\subsection{Multi-subject Interactions}
\subsubsection{Statistic of topic}
balancing(1), tug of warring(1), encouraging(1), synchronizing(1), applauding(1), rowing together(1), spinning(1), jumping together(1), supporting(1), swinging(1), lifting(1), hugging(1), carrying(1), holding hands(1), waving(1), flipping(1), celebrating(1), tossing(1), wrestling(1), stacking(1), whispering(1), cheering(1), gliding(1), leaping(1), skating(1), sliding(1), twirling(1), sharing(1), paddling(1), singing(1), drumming(1), dodging(2), shielding(1), twisting(1), stomping(1), kicking(1), leaning(1), pulling(1), pushing(1), vaulting(1), clashing(1), peering(2), rolling(1), hopping(1), shaking(1), bowing(1), building(1), nesting(1), scurrying(1), foraging(1), weaving(1), diving(1), circling(1), peeking(1), nestling(1), peeling(1), sniffing(1), chirping(1), dashing(1), pouncing(1), snuggling(1), flicking(1), surfing(1), linking(1), bumping(1), jumpstarting(1), guiding(1), swing dancing(1), hovering(1), sledding(1), twinkling(1), zipping(1), balancing on one foot(1), batoning(1), beak tapping(1), blending in(1), blocking(1), blooming together(1), blowing kisses(1), bouncing off(1), bridge building(1), celebratory jumping(1), clambering(1), clapping(1), clasping(1), climbing a rope(1), clockwise spinning(1), coat sharing(1), codebreaking(1), coin flipping(1), colliding(1), composing(1), concocting(1), conga lining(1), contemplating(1), cork popping(1), corn husking(1), counting stars(1), crab walking(1), cracking knuckles(1), creating shadows(1), crisscrossing(1), croquet playing(1), dashing through snow(1), daydreaming(1), defying gravity(1), disappearing act(1), docking(1), dodgeballing(1), dolphin surfing(1), double jumping(1), drag racing(1), ducking(1), echoing(1), egg balancing(1), elbow bumping(1), embracing(1), eye winking(1), face painting(1), fan waving(1), fast forwarding(1), feather tickling(1), fence jumping(1), firework launching(1), fishing together(1), flashlight signaling(1), flipping pages(1), fluttering(1), freezing in place(1), frisbee tossing(1), frog leaping(1), fumbling(1), game playing(1), gazing(1), ghost hunting(1), gift exchanging(1), gliding on air(1), glueing(1), goal scoring(1), gondola riding(1), grappling(1), grinning(1), guarding(1), guessing(1), gymnastics(1), hair braiding(1), hand painting(1), hand shaking(1), hand standing(1), harmonizing(1), harnessing wind(1), head bobbing(1), head butting(1), hearing secrets(1), heel clicking(1), hide and seeking(1), hiking(1), hill rolling(1), home run hitting(1), hopping on one foot(1), horseback riding(1), hula hooping(1), ice skating(1), improvising(1), inventing(1), jigsaw puzzling(1), jumping jacks(1), kicking a can(1), kneeling(1), laughing(1), leaf pile jumping(1), leapfrogging(1), letter writing(1), lifting weights(1), light painting(1), limboing(1), line dancing(1), listening to music(1), log rolling(1), looking through binoculars(1), magician acting(1), mapping(1), meditating(1), metal detecting(1), moon watching(1), moth chasing(1), mountain climbing(1), mushroom picking(1), bouncing(1), huddling for warmth(1), jumping rope(1), high fiving(1), dancing(1), lifting together(1), balancing a ball(1), digging together(1), building a nest(1), sharing food(1)

\subsubsection{Selected prompts}
\begin{adjustwidth}{2em}{0em}
\begin{enumerate}
    \item \textbf{Topic}: Bouncing
    \begin{description}
        \item[\textbf{Prompt}:] A frog and a grasshopper take turns bouncing across lily pads on a pond.
    \end{description}
    \item \textbf{Topic}: Huddling-for-Warmth
    \begin{description}
        \item[\textbf{Prompt}:] A walrus and a polar bear huddle together on an ice floe, sharing warmth.
    \end{description}
    \item \textbf{Topic}: Jumping-Rope
    \begin{description}
        \item[\textbf{Prompt}:] A kangaroo and a lemur each hold an end of a vine, hopping over it together in turn.
    \end{description}
    \item \textbf{Topic}: High-Fiving
    \begin{description}
        \item[\textbf{Prompt}:] Two monkeys jump up and high-five with their paws, celebrating a successful foraging trip.
    \end{description}
    \item \textbf{Topic}: Dancing
    \begin{description}
        \item[\textbf{Prompt}:] Two flamingos perform an elegant dance, mirroring each other's wing movements in perfect coordination.
    \end{description}
    \item \textbf{Topic}: Lifting-Together
    \begin{description}
        \item[\textbf{Prompt}:] Two ants collaborate to lift a large crumb, balancing it carefully as they carry it back to their colony.
    \end{description}
    \item \textbf{Topic}: Balancing-a-Ball
    \begin{description}
        \item[\textbf{Prompt}:] Two seals balance a ball on their noses, passing it back and forth in a coordinated game.
    \end{description}
    \item \textbf{Topic}: Digging-Together
    \begin{description}
        \item[\textbf{Prompt}:] Two meerkats dig a hole side-by-side, their paws flying in rhythm as they excavate a burrow.
    \end{description}
    \item \textbf{Topic}: Building-a-Nest
    \begin{description}
        \item[\textbf{Prompt}:] Two birds bring twigs and leaves to a tree branch, weaving them together to create a shared nest.
    \end{description}
    \item \textbf{Topic}: Sharing-Food
    \begin{description}
        \item[\textbf{Prompt}:] Two bears share a large fish, taking turns taking bites while watching out for other animals.
    \end{description}
\end{enumerate}
\end{adjustwidth}

\subsection{Compositional Spatial Relationships}
\subsubsection{Statistic of topic}
\begin{itemize}
    \item \textbf{Abstract Layouts:}
    chessboard(1), domino(1), constellation(1), pyramid(1), labyrinth(1), kaleidoscope(1), circuit(1), tetris(1), sundial(1), hourglass(1), compass(1), map(1), blueprint(1), gear(1), vortex(1), tessellation(1), barcode(1), spectrum(1), origami(1), satellite(1), silhouette(1), shadow(1), footprint(1), bridge(2), tunnel(1), stained glass(1), windmill(1), lighthouse(1), mountain range(1), river delta(1), waterfall(1), thunderbolt(1), sand dune(1), cliff(1), canyon(1), volcano(2), coral reef(1), aurora(1), nebula(1), eclipse(1), supernova(1), galaxy(1), comet(1), meteor(1), black hole(1), crystal(1), beehive(1), chess knight(1), rubik’s cube(1), sudoku(1), hieroglyph(1), calligraphy(1), musical note(1), soundwave(1), microchip(1), pixel(1), digital clock(1), keyboard(1), mouse pointer(1), barcode scanner(1), jigsaw(1), metro map(1), circuit board(1), telescope(1), microscope(1), hour hand(1), ice crystal(1), tornado(1), tidal wave(1), flame(1), fog(1), reflection(1), horizon(1), globe(1), water cycle(1), ecosystem(1), double helix(1), electric arc(1), solar flare(1), magnet field(1), pendulum(1), gyroscope(1), whirlpool(1), sand timer(1), prism(1), steam power(1), cogwheel(1), marble rolling(1), beam splitter(1), tesseract(1), mobius strip(1), klein bottle(1), electron cloud(1), time lapse(1), solar system(1), tidal force(1), magnetic levitation(1), hologram(1), lens flare(1), binary code(1), algorithm(1), probability tree(1), data cloud(1), social network(1), venn diagram(1), flowchart(1), decision tree(1), optical fiber(1), cosmic web(1), interstellar map(1), dna strand(1), chromosome(1), protein structure(1), enzyme(1), bacteria colony(1), virus model(1), periodic table(1), crystal lattice(1), liquid drop(1), bubble(1), soap film(1), oil slick(1), lava flow(1), fossil(1), seismograph(1), tsunami(1), weather front(1), storm path(1), thundercloud(1), cloud formation(1), rainforest(1), food chain(1), coral polyp(1), ocean current(1), tide pool(1), glacier(1), iceberg(1), volcano cross section(1), fossilized leaf(1), desert oasis(1), lava lamp(1), windmill blades(1), compass needle(1), sundial marks(1), shadow clock(1), metronome(1), pendulum wave(1), ripple tank(1), icicle drip(1), salt crystal(1), gemstone cut(1), light beam bend(1), fiber optic glow(1), laser beam(1), nebula cluster(1), star map(1), supernova explosion(1), gravitational wave(1), celestial sphere(1), solar eclipse(1), lunar cycle(1), moondust(1), martian canyon(1), space dust(1), cosmic string(1), dark matter(1), quark structure(1), higgs boson(1), neutrino path(1), time dilation(1), tic tac toe(1), table(1), atom(1), forest(1), city(1), tree(1), bookshelf(1), flower(1), island(1), garden(1), mosaic(1), 

    \item \textbf{Geometric patterns:}
    snowflake(1), spiral(1), ripple(1), waveform(1), parabola(1), arch(1), infinity symbol(1), yin yang(1), mandala(1), fibonacci sequence(1), sphere(1), cone(1), dodecahedron(1), helix(1), triangle(1), zigzag leaves(1), circle(1)
\end{itemize}
\subsubsection{Selected prompts}
\begin{adjustwidth}{2em}{0em}
\begin{enumerate}
    \item \textbf{Topic}: Zigzag-Leaves
    \begin{description}
        \item[\textbf{Prompt}:] A zigzag pattern made of scattered autumn leaves, creating a path that alternates left and right as it moves forward.
    \end{description}
    \item \textbf{Topic}: Circle
    \begin{description}
        \item[\textbf{Prompt}:] A circle of sunflowers with a single, vibrant red rose in the very center, surrounded by the larger yellow blooms.
    \end{description}
    \item \textbf{Topic}: Tic-Tac-Toe
    \begin{description}
        \item[\textbf{Prompt}:] A tic-tac-toe composed by tomato and cucumber as the players move, and sticks as the grid.
    \end{description}
    \item \textbf{Topic}: Table
    \begin{description}
        \item[\textbf{Prompt}:] A table formed by pretzels stacked together.
    \end{description}
    \item \textbf{Topic}: Atom
    \begin{description}
        \item[\textbf{Prompt}:] An atom depicted with orange as the nucleus and blueberries as electrons spinning in circular orbits.
    \end{description}
    \item \textbf{Topic}: Triangle
    \begin{description}
        \item[\textbf{Prompt}:] A triangle made of sushi pieces, where each side is formed by a different sushi roll.
    \end{description}
    \item \textbf{Topic}: Forest
    \begin{description}
        \item[\textbf{Prompt}:] A forest made from broccoli trees, with animal-shaped cookies as wildlife and a path of cookie crumbs winding through it.
    \end{description}
    \item \textbf{Topic}: City
    \begin{description}
        \item[\textbf{Prompt}:] A cityscape made of stacked crackers as buildings, licorice strips as roads.
    \end{description}
    \item \textbf{Topic}: Tree
    \begin{description}
        \item[\textbf{Prompt}:] A tree made of a pretzel stick as the trunk, with green gummy leaves and woolen yarn roots branching out.
    \end{description}
    \item \textbf{Topic}: Bookshelf
    \begin{description}
        \item[\textbf{Prompt}:] A bookshelf made from colorful candies.
    \end{description}
    \item \textbf{Topic}: Flower
    \begin{description}
        \item[\textbf{Prompt}:] A flower made by colorful gummy.
    \end{description}
    \item \textbf{Topic}: Bridge
    \begin{description}
        \item[\textbf{Prompt}:] A bridge constructed from graham crackers.
    \end{description}
    \item \textbf{Topic}: Island
    \begin{description}
        \item[\textbf{Prompt}:] An island scene with coconut flakes as sand, candy trees on the beach, and blue jelly water as sea.
    \end{description}
    \item \textbf{Topic}: Volcano
    \begin{description}
        \item[\textbf{Prompt}:] A volcano built from chocolate, with red jelly spilling as lava and cotton candy smoke billowing from the top.
    \end{description}
    \item \textbf{Topic}: Garden
    \begin{description}
        \item[\textbf{Prompt}:] A garden made from crushed cookie soil, flower-shaped candies, and wafer cookie paths winding through it.
    \end{description}
\end{enumerate}
\end{adjustwidth}

\end{document}